\documentclass[preprint, 3p]{elsarticle}

\usepackage{easyReview}
\usepackage{amssymb}
\usepackage{amsmath}
\usepackage{lineno}
\usepackage{subcaption}
\usepackage{xcolor}
\usepackage{booktabs}
\PassOptionsToPackage{hyphens}{url}\usepackage{hyperref}
\usepackage{wrapfig}
\newtheorem{definition}{\textbf{Definition}}
\usepackage{float}
\usepackage{algorithm}
\usepackage{algpseudocode}

\begin{document}

\begin{frontmatter}



\title{Safe Continual Reinforcement Learning in Non-stationary Environments}

\author{\large Austin Coursey, Abel Diaz-Gonzalez, Marcos Quinones-Grueiro, and Gautam Biswas \\
Institute for Software Integrated Systems, Vanderbilt University \\
$^*$austin.c.coursey@vanderbilt.edu}
\date{}

\cortext[cor1]{Corresponding author}

%% Abstract
\begin{abstract}
Reinforcement learning (RL) offers a compelling data-driven paradigm for synthesizing controllers for complex systems when accurate physical models are unavailable; however, most existing control-oriented RL methods assume stationarity and, therefore, struggle in real-world \textit{nonstationary} deployments where system dynamics and operating conditions can change unexpectedly. Moreover, RL controllers acting in physical environments must satisfy safety constraints throughout their learning and execution phases, rendering transient violations during adaptation unacceptable. Although continual RL and safe RL have each addressed non-stationarity and safety, respectively, their intersection remains comparatively unexplored, motivating the study of safe continual RL algorithms that can adapt over the system's lifetime while preserving safety. In this work, we systematically investigate \textit{safe continual reinforcement learning} by introducing three benchmark environments that capture safety-critical continual adaptation and by evaluating representative approaches from safe RL, continual RL, and their combinations. Our empirical results reveal a fundamental tension between maintaining safety constraints and preventing catastrophic forgetting under non-stationary dynamics, with existing methods generally failing to achieve both objectives simultaneously. To address this shortcoming, we examine \textit{regularization-based strategies} that partially mitigate this trade-off and characterize their benefits and limitations. Finally, we outline key open challenges and research directions toward developing safe, resilient learning-based controllers capable of sustained autonomous operation in changing environments.
\end{abstract}

%% Keywords

\end{frontmatter}



% \linenumbers

\section{Introduction} \label{sec:introduction}

Reinforcement learning (RL) agents have been successfully applied to challenging autonomous decision-making tasks where sufficiently complete and accurate system models are not available. They have shown strong performance in domains such as robotic control \citep{hwangbo2019learning}, autonomous driving \citep{kiran2021deep}, and drone control \citep{song2023reaching}. However, as RL agents are deployed in real-world settings, they must contend with practical challenges that are often absent or simplified in simulation environments, where they are typically developed and tested \citep{kim2021survey}. Two particularly important challenges are handling non-stationarity \citep{xie2021deep} and satisfying safety constraints during operation \cite{gu2024review}. Many real environments are non-stationary, meaning their dynamics change over time. RL agents operating in such environments must rapidly adapt to these changes, while also retaining knowledge of previously encountered conditions in case they recur. At the same time, real systems must meet safety constraints during both learning and deployment. Thus, an agent must remain safe even as it adapts to changing dynamics.

As a motivating example, consider a Mars rover on a long-duration space mission. Although an RL agent can be trained under conditions that approximate Mars, the true environment will differ due to unseen terrain, weather variability, and the gradual degradation of rover components. An RL agent must, therefore, adapt to evolving, non-stationary dynamics. Moreover, if the rover revisits previously encountered terrain, the agent should not need to relearn how to act effectively in those previously encountered conditions. Simultaneously, the agent’s behavior must comply with safety constraints, as damaging the rover could jeopardize the entire mission. Similar requirements arise in many long-horizon real-world deployments, making this a central problem for real-world control.

To address non-stationarity, the emerging field of \textit{continual reinforcement learning} (CRL) \citep{khetarpal2022towards} studies how agents can learn over a sequence of changing tasks or environments while avoiding \textit{catastrophic forgetting} \citep{kemker2018measuring}. In the rover example, avoiding catastrophic forgetting corresponds to retaining competence on previously experienced terrain types. Separately, the field of \textit{safe RL} has developed methods to ensure agents meet specified safety constraints \citep{brunke2022safe}, such as preventing tip-over events or unsafe maneuvers. Despite substantial progress in both areas, their interaction has received relatively little attention. For example, it remains unclear whether a continual RL agent will reliably remember safety constraints after adapting, or whether a safe RL method will itself suffer from catastrophic forgetting under non-stationarity. Existing work on safe continual RL is limited to a small number of recent studies.

In this study, we investigate the problem of safe continual reinforcement learning and propose two candidate algorithms. We study this problem in the setting of extreme, detectable environment changes with soft safety constraints, controlled by on-policy RL methods capable of real-time adaptation and control. We first define \textit{safe continual RL} problem and further elaborate on the problem setting. We then develop and identify three benchmark environments suitable for evaluating safe continual RL methods, making the intersection of these two fields measurable. Next, we empirically compare several safe RL and continual RL algorithms on both safety and continual learning performance. Finally, we discuss open research questions and future directions for safe continual RL. Code for this paper is available at: \url{https://github.com/MACS-Research-Lab/safe-crl}. 

Overall, our paper makes the following contributions.
\begin{itemize} 
    \item We explore in-depth the problem of safe continual reinforcement learning, providing evidence of its complexity and the inadequacy of current methods. 
    \item We develop and identify three robotic benchmarks for safe continual reinforcement learning: (1) controlling the running of a damaged half-cheetah with velocity constraints, (2) controlling the crawling of a damaged multi-leg ant robot with velocity constraints, and (3) a robotic arm that must avoid spilling a coffee mug while performing everyday tasks like moving an object.
    \item We develop two candidate safe continual RL algorithms called \textit{Safe Elastic Weight Consolidation} (\textbf{Safe EWC}) and \textit{Cost-Fisher Elastic Weight Consolidation} (\textbf{CF EWC}), and demonstrate their potential and limitations as initial solutions to safe continual reinforcement learning. 
    \item We benchmark several safe RL algorithms, several continual RL algorithms, and a simple safe continual RL algorithm using both safe and continual RL metrics. In this process, we identify and discuss several open research questions and directions for safe continual RL. 
    \end{itemize}

The remainder of the paper is structured as follows. Section \ref{sec:literature} describes relevant related work from the continual, safe, and safe continual RL research areas. Section \ref{sec:problem} formally defines the problem of safe continual reinforcement learning and outlines our assumptions about the problem. Section \ref{sec:benchmarks} describes the safe continual RL benchmarks we use. Section \ref{sec:methods} outlines and categorizes the methods we use to explore safe continual RL problems. Section \ref{sec:results} describes our results. Section \ref{sec:discussion} discusses findings and open questions in safe continual RL. Finally, Section \ref{sec:conclusion} concludes the paper. 
\section{Related Work} \label{sec:literature}

In this section, we briefly review the relevant areas of Adaptive Control, Continual RL, Safe RL, and Safe Continual RL.

\noindent
\textbf{Adaptive Control.} Control of systems under non-stationary dynamics has traditionally been solved using adaptive control methods \citep{bar1988simple}. In adaptive control, the controller's parameters are adjusted online to compensate for changes or uncertainties in the system's dynamics. It has been used with great success for provably optimal control in non-stationary environments. These include applications, such as quadrotor control under parametric uncertainties \citep{dydek2012adaptive} and additive manufacturing \citep{xiong2014adaptive}. Traditional adaptive control approaches, like Model Reference Adaptive Control \cite{nguyen2018model}, require a known model structure and often linearly parameterize the unknown non-stationarity. This can limit the applicability of these methods, as many real systems lack known models and real non-stationarity is not always linear. 

With this in mind, Model-Free Adaptive Control \citep{hou2013model} methods, such as a pseudo-partial derivative dynamic linearization technique \citep{hou2011data} and a multiple adaptive observer technique \citep{xu2014novel}, have been developed. These methods leverage data to approximate the system's behavior in an adaptive control scheme. This has been used in domains such as fixed-wing drone control in windy conditions \citep{zhao2017model} and cyber-physical systems under cyber attacks \citep{ma2022dynamic}. While these methods can avoid the need for accurate system models, they lack mechanisms for remembering. In other words, if previously-seen dynamics are revisited, they will typically need to relearn their model. To avoid relearning previously seen dynamics without a system model while adapting to non-linear, non-stationary environments, we can turn toward continual reinforcement learning.

\noindent
\textbf{Continual Reinforcement Learning.} Continual reinforcement learning algorithms operate in a non-stationary environment and must balance the stability and plasticity of their underlying networks \citep{khetarpal2022towards}. The \textit{stability} of a network is defined by its ability to retain previous knowledge, such as its performance in previous environmental conditions in a non-stationary environment. The \textit{plasticity} of a network is its flexibility to adapt to and learn new information. Many recent works study or aim to reduce plasticity loss while learning reinforcement learning agents \citep{klein2024plasticity, abbas2023loss, nikishin2023deep, juliani2025study}. 

Aside from plasticity loss, recent work on continual reinforcement learning has largely focused on developing techniques to improve stability. In particular, avoiding catastrophic forgetting is a central problem for continual reinforcement learning research \citep{khetarpal2022towards}. Informally, a reinforcement learning agent catastrophically forgets when it fails to perform as well as it did previously on a task. Techniques to reduce forgetting can be divided into regularization  \citep{kaplanis2018continual, kirkpatrick2017overcoming, schwarz2018progress}, replay \citep{rolnick2019experience, pathmanathan2024using, li2021sler}, and expansion-based techniques  \citep{kessler2020unclear, zhang2023dynamics, xu2020task}. 

Regularization-based approaches, such as EWC \citep{kirkpatrick2017overcoming}, reduce forgetting in neural networks by penalizing changes to previously important weights. Replay-based approaches use replay buffers to store information from prior tasks. When the policy is updated, recent observations and historical data can be used to ensure the model balances current performance with forgetting. This replay buffer can be explicit \citep{rolnick2019experience} or generated \citep{li2021sler}. Finally, expansion techniques mitigate forgetting by expanding parts of the network, such as the mixture model in \cite{xu2020task}. While research on the stability-plasticity dilemma and catastrophic forgetting is important for lifelong reinforcement learning, it does not address how to satisfy safety constraints, a key aspect of real systems. 

\noindent
\textbf{Safe Reinforcement Learning.} The study of safety constraints within reinforcement learning has been comprehensively explored in the context of safe reinforcement learning \citep{brunke2022safe}. Various methodologies have been employed to enforce these safety constraints, including primal \citep{xu2021crpo} and primal-dual \citep{liu2022constrained} optimizations of constrained Markov decision processes, along with other techniques such as Lyapunov functions \citep{zhang2021safe}, control barrier functions \citep{emam2022safe}, implicit learning \citep{srinivasan2020learning}, model predictive control \citep{lutjens2019safe}, and reachability analysis \citep{selim2022safe}. 

Generally, safe reinforcement learning methods assume fixed environments, which do not adequately account for the non-stationarity inherent in long-term real-world deployments. Nonetheless, some research has delved into safety in non-stationary contexts \citep{chandak2020towards, chen2021context, ding2023provably}; however, these studies primarily focus on safety in general non-stationary environments, overlooking the unique challenges associated with continual or lifelong reinforcement learning.

\noindent
\textbf{Safe and Continual Reinforcement Learning.} From the above literature review, it is clear that there is a research gap in safe, continual reinforcement learning methods. While there is some work on safe reinforcement learning in related domains such as meta-learning \citep{cho2024constrained} and transfer learning \citep{zhang2024safety, wang2024transfer}, there is little work on safety in a continual reinforcement learning setting. The first safe, continual RL method was developed in \cite{ammar2015safecrl}, where the authors learned safe control policies across multiple simple tasks, without evaluating metrics, such as forgetting and transfer. More recently, in \cite{ganie2024safecrl}, the authors developed a safe continual reinforcement learning technique using barrier functions, layer regularization, and new weight updates for mobile robots. However, their approach is validated only in limited environments, without safe or continuous learning metrics. In our previous work \citep{coursey2025safecrl}, we conducted an initial exploration of safe continual RL, but our comparisons were limited to one safe and one continual RL algorithm across environments with a single safety constraint. Given the limitations of current research, this paper conducts a deep exploration of the problem of safe continual reinforcement learning.

\section{Defining Safe Continual Reinforcement Learning} \label{sec:problem}

Before formally defining a safe continual reinforcement learning problem, we clarify our focus and assumptions about the nature of the problem.

\begin{itemize}
    \item \textbf{Discrete, dramatic task changes.} We assume environmental changes occur as discrete tasks. While real non-stationarity can be gradual and subtle, our interest is in more challenging cases involving abrupt, substantial changes in dynamics or conditions.
    \item \textbf{Known task boundaries.} We assume task changes are known to the agent (i.e., task boundaries are given). Without this information, continual RL performance would be limited by the quality of a separate task identification mechanism. Our goal is to evaluate continual RL methods themselves rather than task inference, and given the dramatic task shifts we consider, this assumption is reasonable. 
    \item \textbf{Soft safety constraints.} We model safety as a soft constraint, meaning safety is enforced in expectation (or via penalties/constraint satisfaction on average) rather than as a hard requirement that must never be violated. This allows us to study the trade-off between safety and continual adaptation. In contrast, hard safety constraints would require the agent to never violate safety, making ``never forgetting” safety-critical behavior essential and placing strict demands on continual learning.
    \item \textbf{Catastrophic forgetting as the primary continual learning challenge.} We focus on catastrophic forgetting in the continual RL setting: the tendency of agents to lose performance on earlier tasks after learning new ones. Although other aspects of continual RL are important, we view avoiding catastrophic forgetting as a necessary first step toward robust, safe continual learning.
\end{itemize}

To formally define safe continual RL, we start with the definition of a reinforcement learning problem. \textbf{Reinforcement learning} solves a Markov Decision Process (MDP):
\[\text{MDP}=\left(\mathcal{S}, \mathcal{A}, p, r\right),\] 
which is a four-tuple consisting of a state space $\mathcal{S}$, an action space $\mathcal{A}$, $p(s'|s,a)$ is the probability of reaching state $s'$ while performing action $a$ in state $s$, and $r(s,a)$ is the scalar reward obtained by taking action $a$ in state $s$. The RL algorithm is designed to find a policy $\pi(a|s)$ that maximizes the expected discounted reward $V_{\pi}$, for a certain discount factor $\gamma \in [0, 1)$,
\begin{align*}
    V_{\pi}(s)=\mathbb{E}\left[\left.\sum_{i=0}^\infty\gamma^{i}r(s_i,a_i) \right| s_0=s, a_i \sim \pi(\ \cdot\ |s_i), s_{i+1}\sim p(\ \cdot\ |s_i,a_i)\right],\quad  s \in\mathcal{S}.
\end{align*}

\textbf{Safe Reinforcement Learning} is typically formulated as a Constrained Markov Decision Process \citep{altman1999constrained}: 
\[ \text{CMPD} = (\mathcal{S}, \mathcal{A}, p,r, \mathcal{C}, d), \]
where $\mathcal{C}: \mathcal{S} \times \mathcal{A} \to \mathbb{R}^m$ is the vector-valued cost function that determines the $m$ cost-values of the states
\begin{equation*}
    V^{\mathcal{C}_j}_{\pi}(s)=\mathbb{E}\left[\left.\sum_{i=0}^\infty \gamma^i \mathcal{C}_{j}(s_{i}, a_{i})\right| s_0=s, a_i \sim \pi(\ \cdot\ |s_i), s_{i+1}\sim p(\ \cdot\ |s_i,a_i)\right], \quad  s \in\mathcal{S}, \; j=1,2,\dots, m,
\end{equation*}
which are constrained by $m$ maximum costs $d=(d_1, d_2, \dots, d_m)\in \mathbb{R}^m$ 
\begin{equation*}
    V^{\mathcal{C}_j}_{\pi}(s) \leq d_j,\quad  s \in\mathcal{S},\;  j=1,2,\dots, m.
\end{equation*}

\textbf{Continual Reinforcement Learning} (CRL) problem defines an RL problem where ``the best agents never stop learning'' \citep{abel2023definition}. CRL has been recently rigorously defined \citep{abel2023definition}. In this paper, we retain the characteristics of the original CRL problem, but for simplicity, we adopt the framework of a Non-Stationary Markov Decision Process (NSMDP) \citep{LecarpentierR19}.  

\noindent
We consider a 
\[ \text{NSMDP}=\left(\mathcal{S}, \mathcal{A}, \{p^{(\tau)}\}_{\tau\in\mathcal{T}}, \{r^{(\tau)}\}_{\tau\in\mathcal{T}}\right), \]
as an extension of a stationary MDP, where the transition probability and reward are functions of a set of environmental conditions, also called tasks, $\mathcal{T}\subset\mathbb{N}$. Our goal is to find a policy that learns tasks consecutively, acquiring multiple trajectories within each task before moving to the next, with no control over the task order, and that can quickly adapt and build on previously learned tasks \citep{khetarpal2022towards}. 

Additionally, a successful continual learning agent must be resistant to catastrophic forgetting, so we impose the following condition on the current policy. 

\noindent
 Let $\mathcal{T}_{\prec}\subset \mathcal{T}$ be the set of tasks that have already been trained on and let $\pi_\kappa^*$ be the optimal policy obtained at task $\kappa\in \mathcal{T}_{\prec}$. Then, for some tolerance to forgetting value $\epsilon$, our current policy  $\pi$ should satisfy
\begin{equation*}
    V^{(\kappa)}_{\pi}(s) \ge V_{\pi^*_\kappa}^{(\kappa)}(s) - \epsilon , \quad s \in\mathcal{S},\; \kappa \in \mathcal{T}_{\prec},
\end{equation*}
where $V_{\pi}^{(\kappa)}$ denotes the expected discounted reward of the policy $\pi$ in the stationary $\text{MDP}=\left(\mathcal{S}, \mathcal{A}, p^{(\kappa)}, r^{(\kappa)}\right)$.

\begin{definition}[Safe Continual Reinforcement Learning] 
\phantomsection
\label{def:safe-crl}
A \textit{safe continual reinforcement learning} problem solves a ``non-stationary constrained Markov decision process''  
\[
\text{NSCMDP}=\left(\mathcal{S}, \mathcal{A}, \{p^{(\tau)}\}_{\tau\in\mathcal{T}}, \{r^{(\tau)}\}_{\tau\in\mathcal{T}},\{\mathcal{C}^{(\tau)}\}_{\tau\in\mathcal{T}},\{d^{(\tau)}\}_{\tau\in\mathcal{T}}\right),
\]
where at each epoch belonging to task $\tau$, we aim to find a policy $\pi^*$ that addresses both continual and safety constraints simultaneously:
\begin{align} \label{eq:safe-crl}
    \pi^*(s) &= \arg\max _{\pi}  V_{\pi}^{(\tau)}(s) \notag \\
    \text{s.t.} \ & \left\{
    \begin{aligned}
        & V^{(\kappa)}_{\pi}(s) \ge V_{\pi^*_\kappa}^{(\kappa)}(s) - \epsilon,  
        \quad  s \in\mathcal{S}, \; \kappa \in \mathcal{T}_{\prec}, \\
        & V^{\mathcal{C}_j,(\kappa)}_{\pi}(s) \leq d^{(\kappa)}_j,  
        \quad  s \in\mathcal{S}, \; \kappa \in \mathcal{T}_{\prec}\cup\{\tau\}, \; j = 1,2,\dots, m_{\kappa},
    \end{aligned} \right.
\end{align}
where $V^{\mathcal{C}_j,(\kappa)}_{\pi}$ denotes the $j$-th cost-value following policy $\pi$ in the stationary $\text{CMDP}=\left(\mathcal{S}, \mathcal{A}, p^{(\kappa)}, r^{(\kappa)},\mathcal{C}^{(\kappa)},d^{(\kappa)}\right)$.
\end{definition}

\section{Benchmark Environments for Safe Continual Reinforcement Learning} \label{sec:benchmarks}

% Describe the two benchmarks we introduce. Mention one is a safe RL benchmark and the other is a continual RL benchmark. Connect the halfcheetah to the ECC paper under submission.

\begin{figure}[t]
    \centering
    \includegraphics[width=\columnwidth]{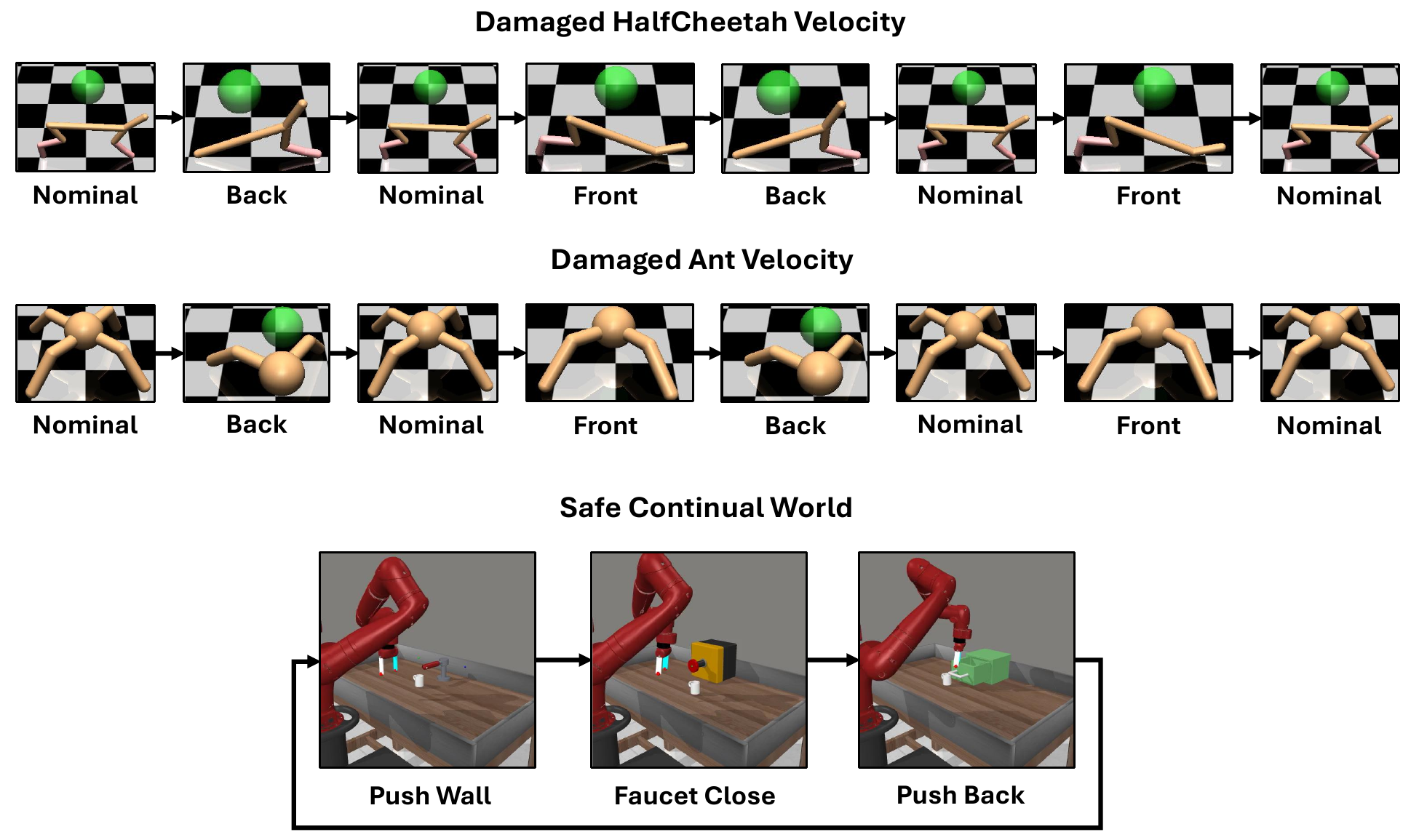}
    \caption{Safe continual reinforcement learning benchmarks. \textbf{Top}: HalfCheetah MuJoCo task with a maximum velocity constraint. Every 1 million timesteps, the cheetah is damaged or repaired, significantly changing its dynamics. \textbf{Middle}: Ant MuJoCo task with a maximum velocity constraint. Every 1 million timesteps, the ant is damaged or repaired, significantly changing its dynamics. \textbf{Bottom}: three tasks from the Safe Continual World benchmark. A coffee mug is placed in the scene, and the agent is penalized for tipping the mug while solving the task. Every 2 million timesteps, the task changes, and the task sequence is repeated once.}
    \label{fig:benchmarks}
\end{figure}

We need suitable benchmarks to investigate and compare different algorithms for safe continuous RL. This paper introduces three robotics benchmarks. The first is a modified version of the classic HalfCheetah MuJoCo environment, which we call the \textit{Damaged HalfCheetah Velocity} environment. The second modifies the Ant MuJoCo navigation environment, referred to as the \textit{Damaged Ant Velocity}. The third benchmark is a modification of the Continual World environment, known as the \textit{Safe Continual World}. 
The first two environments are widely used benchmarks for safe reinforcement learning and have been extended to incorporate non-stationary dynamics. The Safe Continual World environment is a continual reinforcement learning benchmark that incorporates safety constraints. The code for these proposed benchmarks is available on our GitHub repository \url{https://github.com/MACS-Research-Lab/safe-crl}. These environments, shown in Figure \ref{fig:benchmarks}, are described in more detail below.

\subsection{Damaged HalfCheetah Velocity Environment}

The Damaged HalfCheetah Velocity environment extends the popular HalfCheetah Velocity safe RL benchmark \citep{ji2023safety}. In this environment, the agent's goal is to learn to run at a velocity below a constraint by moving the joints of a robotic cheetah. The primary challenge in this environment is balancing the opposing objectives of safety and reward. The HalfCheetah Velocity task lacks non-stationarity and does not represent a safe continual reinforcement learning problem. Some prior work introduces non-stationarity into lifelong learning of the HalfCheetah, including by changing the target velocity/wind \citep{xie2021lilac} or by introducing obstacles in the environment \citep{lee2023skilldiscovery}. In the Damaged HalfCheetah Velocity benchmark, which we proposed in our previous paper \citep{coursey2025safecrl}, the robot is damaged and subsequently repaired, leading to significant changes in its dynamics. The robot cycles through the task sequence \{nominal, back, nominal, front, back, nominal, front, nominal\} with task changes every 1 million timesteps, where nominal is the typical HalfCheetah Velocity configuration, back omits the back leg, and front omits the front leg. This task sequence is shown at the top of Figure \ref{fig:benchmarks}.

Details on the state and action spaces, rewards, costs, and the impact of removing limbs are provided below. Modified from \cite{ji2023safety}, this benchmark uses the standard Gymnasium MuJoCo Half Cheetah environment \footnote{\url{https://gymnasium.farama.org/environments/mujoco/half_cheetah/}} as a base. 

\noindent
\textbf{State Space.} 17-dimensional vector: the z-coordinate of the front tip, the angles of the front tip, back thigh, back shin, back foot, front thigh, front shin, and front foot, the velocity of the z- and x-coordinates of the front tip, and the angular velocities of the angles. 

\noindent
\textbf{Action Space.} 6-dimensional vector controlling the torque $\in[0, 1]$ applied to the rotors on each joint.

\noindent
\textbf{Damaged Scenario.} We implement the damaged scenarios by removing an entire limb from the MuJoCo XML file. The observations for that limb are therefore gone, and the actions that would operate on those joints no longer work (though the size of the action vector does not change). 

\noindent
\textbf{Reward.} The agent is rewarded for moving eastward (increasing $x$) without large actions:

\begin{equation*}
    r_{\text{HalfCheetah}}=\frac{x_{t+1} -x_t}{dt} - 0.1 ||action||_2^2.
\end{equation*}

\noindent
\textbf{Cost.} The agent is penalized for violating the velocity constraint (3.2096 m/s):

\begin{equation*}
    cost = \mathbb{I}(v > 3.2096).
\end{equation*}

\subsection{Damaged Ant Velocity Environment}

This environment extends the popular Ant Velocity safe RL benchmark \citep{ji2023safety}. Similar to the HalfCheetah, it has been modified to introduce non-stationary dynamics by simulating limbs being damaged and repaired at different points in time as the ant operates. Since the ant has four limbs, the two front and two back limbs are removed in the damaged state to destabilize the robot and force the policy to relearn. The ant cycles through the task sequence \{nominal, back, nominal, front, back, nominal, front, nominal\} with task changes occurring every 1 million timesteps, where nominal is the typical Ant Velocity configuration, back is missing the two back legs, and front is missing the two front legs. This task sequence is shown in the middle of Figure \ref{fig:benchmarks}.

Details on the state, action, reward, and cost spaces are provided below. Modified from \cite{ji2023safety}, this benchmark uses the standard Gymnasium MuJoCo Ant environment \footnote{\url{https://gymnasium.farama.org/environments/mujoco/ant/}} as a base. 

\noindent
\textbf{State Space.} 105-dimensional vector: the positions of the ant's body parts (13 elements), the velocities of the body parts (14 elements), and the center of mass based external forces on the body parts (78 elements).

\noindent
\textbf{Action Space.} 8-dimensional vector controlling the torque $\in[-1, 1]$ applied at the hinge joints.

\noindent
\textbf{Reward.} The agent is rewarded for moving eastward (increasing $x$) without large actions, falling over, or having too large of contact forces:

\begin{equation*}
    r_{\text{Ant}}=r_\text{healthy}+\frac{x_{t+1} -x_t}{dt} - 0.5 ||action||_2^2 - 5\times10^{-4}||F_{contact}||_2^2,
\end{equation*}

where $r_\text{healthy}$ is a fixed value reward (1) given to the ant for not terminating the episode by falling over or jumping too high.

\noindent
\textbf{Cost.} The agent is penalized for violating the velocity constraint (2.6222 m/s):

\begin{equation*}
    cost = \mathbb{I}(v > 2.6222).
\end{equation*}

\subsubsection{Safety and Reward Tradeoff}

In the MuJoCo Damaged Velocity environments, the safety constraint is a velocity constraint that remains consistent across tasks. Since the agent's goal is to run as far east as possible, it must learn to increase its eastward velocity. Therefore, a poor policy usually does not violate safety. This creates an interesting scenario with a direct tradeoff between safety and reward.

\subsection{Safe Continual World}

The Safe Continual World environment builds on the Continual World environment \citep{wolczyk2021continual}, which in turn extends the Meta World environment \citep{yu2020meta}. In Meta World, an agent learns to control a robotic arm equipped with a gripper to perform everyday tasks. The Continual World introduces a challenging sequence of tasks focused on issues of forgetting and forward transfer for Meta World. This primary task sequence in Continual World consists of 10 tasks, and it was specifically designed to be demanding, with an estimated execution time of around 100 hours on an 8-core machine \citep{wolczyk2021continual}. 

During our initial experiments, we discovered that most tasks in this sequence could not be reliably solved by on-policy reinforcement learning (RL) methods within the provided timeframe, even without the additional difficulties posed by non-stationary task changes or safety constraints. To simplify the problem and facilitate a deeper study of safe continual reinforcement learning, we defined a new task sequence. This sequence was chosen as a subset of tasks that recent findings indicate can be successfully completed within 3 million timesteps using standard RL methods (see Figure 9 in \citep{ahn2025prevalence}).

Our task sequence is \{faucet-close, button-push, drawer-close\}, repeated once. The goals of these tasks are as follows. Faucet-close: turn a faucet off, button-push: push a button, drawer-close: close an open drawer. All three tasks require the robot to move its arm forward and push an object. The similarity of these tasks is designed to enable RL agents to transfer knowledge across tasks, facilitating the study of both forward and backward transfer. We introduce safety constraints into this non-stationary environment by adding a coffee mug to the table, shown at the bottom of Figure \ref{fig:benchmarks}. This coffee mug is between the robotic arm and the current task's goal. The agent should not spill the coffee while performing its task. To model this constraint, we define cost as the mug's current tilt (in degrees). The mug's location is random, but the agent observes it. This can be thought of as an obstacle avoidance constraint.

Details on the state, action, reward, and cost spaces are provided below. We modify the Continual World \citep{wolczyk2021continual} benchmark to introduce safety constraints. However, since Continual World is just a sequenced version of Meta World \citep{yu2020meta}, we modify Meta World instead. We choose to do this because, at the time of writing, Meta World is actively maintained and updated while Continual World is not.

\noindent
\textbf{State Space.} The Safe Continual World state space is a 53-dimensional vector consisting of the current positions, previous positions, and goal position. The positions are the gripper position, the amount the gripper is open, the position and quaternion of up to two objects relevant to the task (the second is masked if there is only one), and the position and quaternion of the mug that should not be tipped. Each object and goal is initialized in a random position in each episode, and episodes are 500 timesteps long.

\noindent
\textbf{Action Space.} The action space is the desired $[\Delta x, \Delta y, \Delta z, \Delta \text{gripper close}]$ of the robotic arm.

\noindent
\textbf{Reward.} The reward function in Meta World is a dense reward that guides the agent toward completing each step of the task. The reward function is task-dependent. See the Appendix of Meta World \citep{yu2020meta} for details.

\noindent
\textbf{Cost.} In a Safe Continual World, the agent must avoid tipping over the coffee mug while completing its tasks. As a result, the cost function is designed to penalize the agent based on the angle, in degrees, at which the mug is tipped. If the mug is flat on the table, the cost is set at 90. Since this is considered a soft safety constraint, the episode does not end when the mug tips over. In theory, the agent can reorient the tipped mug to reduce the cost.

\begin{algorithm}[!t]
\caption{Safe Elastic Weight Consolidation (Safe EWC)}
\label{alg:safe-ewc}
\begin{algorithmic}[1]
\Require Tasks $\{\mathcal{T}_1,\dots,\mathcal{T}_K\}$
\Require PPO params $(\gamma,\lambda_{\text{GAE}},\epsilon,c_v,c_e)$
\Require Learning rates $(\alpha_\theta,\alpha_\phi)$, EWC coeff.\ $\lambda$, cost coeff.\ $\beta$
\Require Task length $T_{\text{task}}$, update interval $T_{\text{update}}$, Fisher samples $N$

\State Initialize $\pi_\theta$, $V_\phi$, $\mathcal{M}\leftarrow\emptyset$
    \Comment{Policy, value function, and EWC memory}

\For{$k=1$ to $K$}
    \State Initialize environment $\mathcal{T}_k$
        \Comment{Start new task}
    \State $\mathcal{D}\leftarrow\emptyset$
        \Comment{Clear rollout buffer}

    \For{$t=1$ to $T_{\text{task}}$}
        \State $a_t \sim \pi_\theta(\cdot\mid s_t)$
            \Comment{Sample action}
        \State $(s_{t+1}, r_t, C_t)\sim\mathcal{T}_k(s_t,a_t)$
            \Comment{Environment transition}
        \State $\tilde r_t \leftarrow r_t - \beta C_t$
            \Comment{Cost-shaped reward (Safe EWC)}
        \State $\mathcal{D}\leftarrow\mathcal{D}\cup(s_t,a_t,\tilde r_t,s_{t+1})$
            \Comment{Store transition}

        \If{$t \bmod T_{\text{update}} = 0$}

            \State \textbf{PPO return and advantage computation}
            \vspace{0.2cm}
            
            \State $\delta_t \leftarrow \tilde r_t + \gamma V_\phi(s_{t+1}) - V_\phi(s_t)$
                \Comment{TD residual}
            \State $\hat A_t \leftarrow \sum_l (\gamma\lambda_{\text{GAE}})^l\delta_{t+l}$
                \Comment{Generalized advantage estimation (GAE)}
            \State $G_t \leftarrow \sum_l \gamma^l\tilde r_{t+l}$
                \Comment{Discounted returns}
            \vspace{0.2cm}
            \State \textbf{PPO policy and value losses}
            \vspace{0.2cm}
            \State $r_t(\theta)\leftarrow
            \pi_\theta(a_t|s_t)/\pi_{\theta_{\text{old}}}(a_t|s_t)$
                \Comment{Importance ratio}
            \State $\mathcal{L}_{\text{PPO}}\leftarrow
            -\mathbb{E}[\min(r_t\hat A_t,\text{clip}(r_t,1\pm\epsilon)\hat A_t)]$
                \Comment{Clipped surrogate objective}
            \State $\mathcal{L}_{\text{value}}\leftarrow
            \mathbb{E}[(V_\phi(s_t)-G_t)^2]$
                \Comment{Value regression}
            \State $\mathcal{L}_{\text{entropy}}\leftarrow
            -\mathbb{E}[\mathcal{H}(\pi_\theta(\cdot|s_t))]$
                \Comment{Entropy regularization}

            \vspace{0.2cm}
            \State \textbf{Elastic Weight Consolidation (EWC)}
            \vspace{0.2cm}
            
            \State $\mathcal{L}_{\text{EWC}}\leftarrow
            \sum_{(\theta^*_j,F_j)\in\mathcal{M}}
            \sum_i \frac{\lambda}{2}F_{j,i}(\theta_i-\theta^*_{j,i})^2$
                \Comment{Penalty for forgetting previous tasks}

            \vspace{0.2cm}
            \State \textbf{Parameter updates}
            \vspace{0.2cm}
            
            \State $\mathcal{L}_\theta\leftarrow
            \mathcal{L}_{\text{PPO}}+c_e\mathcal{L}_{\text{entropy}}+\mathcal{L}_{\text{EWC}}$
                \Comment{Total actor loss}
            \State $\mathcal{L}_\phi\leftarrow c_v\mathcal{L}_{\text{value}}$
                \Comment{Total critic loss}
            \State $\theta\leftarrow\theta-\alpha_\theta\nabla_\theta\mathcal{L}_\theta$
                \Comment{Update policy}
            \State $\phi\leftarrow\phi-\alpha_\phi\nabla_\phi\mathcal{L}_\phi$
                \Comment{Update value function}

            \State $\theta_{\text{old}}\leftarrow\theta,\ \mathcal{D}\leftarrow\emptyset$
                \Comment{Prepare next rollout}
        \EndIf
    \EndFor
    
    \State $\theta^*_k\leftarrow\theta$
        \Comment{Save task-optimal parameters}
    \State Sample $N$ transitions $\{(s_n, a_n)\}_{n=1}^N$ from $\mathcal{T}_k$ using $\pi_{\theta^*_k}$
        \Comment{Fisher rollout}
    \State $F_{k,i}\leftarrow
    \frac{1}{N}\sum_n(\partial_{\theta_i}\log\pi_{\theta^*_k}(a_n|s_n))^2$
        \Comment{Estimate Fisher information}
    \State $\mathcal{M}\leftarrow\mathcal{M}\cup(\theta^*_k,F_k)$
        \Comment{Store task importance}
\EndFor
\end{algorithmic}
\end{algorithm}

\section{Safe and Continual Reinforcement Learning Methods} \label{sec:methods}
In this section, we present safe continual RL algorithms based on elastic weight consolidation (EWC) \citep{kirkpatrick2017overcoming} as the foundational continual learning mechanism. Two algorithms, (1) Safe Elastic Weight Consolidation (Safe EWC) and (2) Cost-Fisher Elastic Weight Consolidation (CF-EWC), are described in greater detail after introducing the core framework of our safe continual RL approach.

\subsection{Proposed Safe Continual Reinforcement Learning Algorithms}

To address the safe continual reinforcement learning problem, we propose two new algorithms. These algorithms build on the PPO+EWC framework \citep{nath2023modulating} by incorporating safety constraints. We chose elastic weight consolidation (EWC) \citep{kirkpatrick2017overcoming} as the foundational continual learning mechanism because of its simplicity and widespread use. EWC helps mitigate the forgetting problem in continual learning by penalizing significant changes to parameters that were important in previous tasks, effectively ``freezing” certain parts of the network. The loss regularization is formulated as follows:

\begin{equation} \label{eq:ewc}
    \mathcal{L}(\theta) = \mathcal{L}_B(\theta) + \sum_i\frac{\lambda}{2}F_i(\theta_i-\theta^*_{A,i})^2,
\end{equation}

\noindent
where $\mathcal{L}_B$ is the loss on the current task, $F$ is the Fisher information matrix measuring the ``importance'' of each parameter, $\lambda$ is a hyperparameter to balance current and previous tasks, and $\theta_A$ are parameters for a previous task. In practice, this diagonal of $F$ is approximated using the following equation:

\begin{equation} \label{eq:fisher}
    F_i \approx \frac{1}{N} \sum_{n=1}^{N} \left( \frac{\partial}{\partial \theta_i} \log p(y_n | x_n; \theta^*_A) \right)^2,
\end{equation}

\noindent
where $N$ is the number of samples in a dataset from prior task $A$, $\theta^*_A$ are the saved weights from the network learned on prior task $A$, and $p(y_n | x_n; \theta^*_A)$ is the probability of the output $y_n$ given the input and weights (the policy/actor network). So, for each weight in the saved parameters, its importance across a representative dataset from the previous task is averaged. This is then used to regularize the (policy) network loss as shown in Equation \eqref{eq:ewc}.

\subsubsection{Algorithm 1: Safe Elastic Weight Consolidation (Safe EWC)} \label{sec:safe-ewc}

We extend PPO+EWC by modifying the loss in Equation \eqref{eq:ewc} to incorporate safety constraints, yielding Safe EWC, as introduced in prior work \citep{coursey2025safecrl}. This is achieved by shaping the reward with the cost; specifically, we define the modified reward as follows:

\begin{equation} \label{eq:safe-ewc}
    r_{\text{Safe EWC}}(s, a) = r(s, a) - \beta C(s, a),
\end{equation}

\noindent
where $\beta$ is a hyperparameter that determines the cost/reward tradeoff and $C:S\times A\to\mathbb{R}^m_{+}$ is the current task cost function. In our experiments, we set $\beta=5$, which induces a direct trade-off between safety and reward. An optimal policy maximizes reward while incurring no safety violations. This shaping implicitly mitigates forgetting of safety constraints by encoding them in the reward signal; however, it modifies the reward function and may adversely affect learning in complex tasks with carefully designed rewards.

Algorithm \ref{alg:safe-ewc} details our Safe EWC  method. We present the proximal policy optimization (PPO) computations to illustrate how elastic weight consolidation (EWC) regularizes the actor loss; for PPO components, see \citep{schulman2017proximal}. Our implementation, adapted from \citep{ji2023safety}, includes standard PPO stabilization techniques, including critic regularization. We also maintain a cost critic to monitor safety signals, although policy optimization uses only cost-shaped rewards. The algorithm iterates over the task sequence; however, in real settings, task identities are often unknown, and task changes must be detected. During each task, the agent is trained with PPO using reward shaping from Equation \eqref{eq:safe-ewc} (line 8). After the first task, a representative dataset is sampled to compute task-specific Fisher information, capturing parameter importance (line 32); we use the final episode as this dataset. For subsequent tasks, the actor updates apply Equations \eqref{eq:fisher} and \eqref{eq:ewc} to penalize changes to parameters deemed important for previous tasks (line 21). 

\subsubsection{Algorithm 2: Cost-Fisher Elastic Weight Consolidation (CF-EWC)} \label{sec:cf-ewc}

To address safe continual reinforcement learning without modifying the reward function, we propose a second EWC extension, Cost-Fisher Elastic Weight Consolidation (CF-EWC). In this algorithm, we modify the computation of the approximated Fisher information matrix to be:

\begin{equation} \label{eq:cf-ewc}
    F_i \approx \frac{1}{N} \sum_{n=1}^{N} \frac{1}{(c_n+1)}\left( \frac{\partial}{\partial \theta_i} \log p(y_n | x_n; \theta^*_A) \right)^2,
\end{equation}

\noindent
where $c_n$ is the positive cost of sample $n$ and the rest is the same as Equation \eqref{eq:fisher}. Intuitively, this assigns higher importance to the weights with higher probabilities that had lower costs. When there is no cost, this computes the typical Fisher information. When the cost is high, the weight of the parameters' importance shrinks toward zero. Therefore, we are allowing the agent to modify the weights that violated safety on previous tasks, while discouraging modifications that had a lower cost with a high reward. Note that if the cost scale changes over time (as allowed by Definition \ref{def:safe-crl}), this cost may need to be normalized. 

The complete algorithm is omitted from the main text for brevity and presented in the Appendix as Algorithm \ref{alg:cf-ewc}. The only substantive differences from Algorithm \ref{alg:safe-ewc} are the use of Equation \ref{eq:cf-ewc} in line 32 and the removal of cost-based reward shaping from line 8.

\subsection{Safe or Continual Reinforcement Learning Baselines}

\begin{wrapfigure}{r}{0.35\textwidth}
    \centering
    \includegraphics[width=\linewidth]{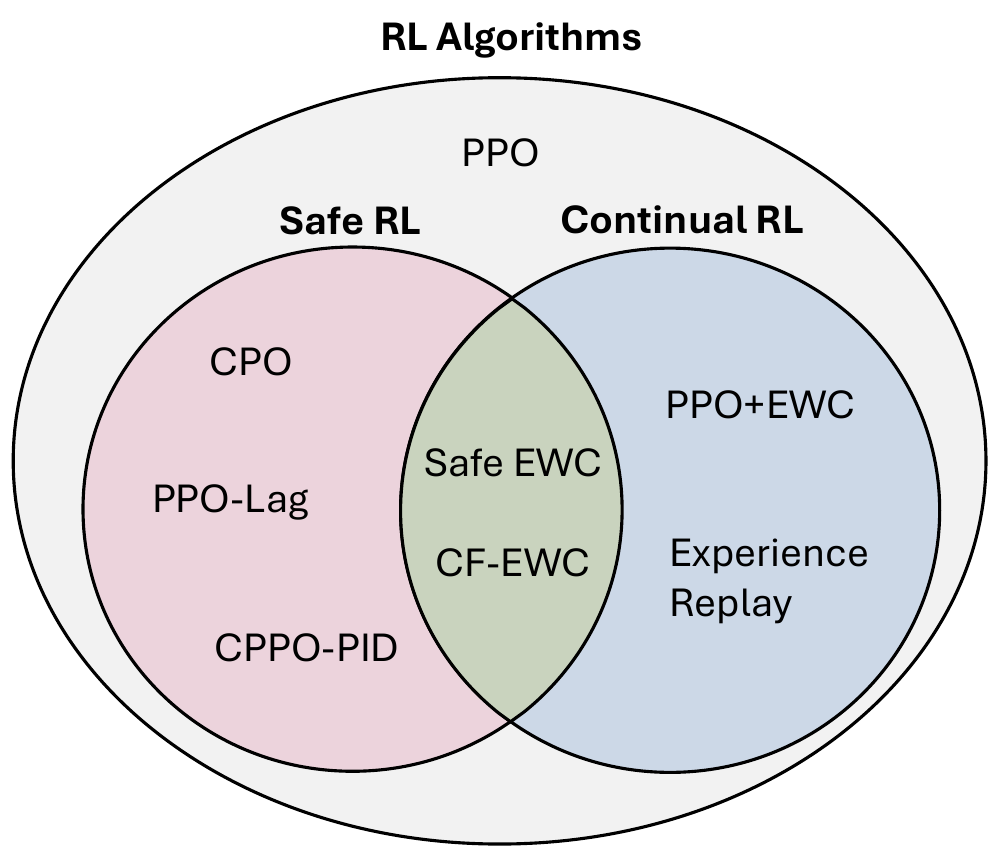}
    \caption{Categorization of RL algorithms used in this study. Safe continual RL algorithms, shown in the green intersection, should satisfy the requirements of both safe and continual RL.}
    \label{fig:algorithms}
\end{wrapfigure}

This paper aims to enhance understanding of safe continual reinforcement learning by identifying key challenges and open questions, thereby motivating the development of frameworks that integrate safe and continual RL to address them. We benchmark safe RL, continual RL, safe continual RL, and standard RL methods on the environments described in Section \ref{sec:benchmarks}. The selected methods are widely used in their respective subfields. All methods employ trust-region, on-policy policy-gradient algorithms (PPO or TRPO) \citep{schulman2015trust,schulman2017proximal}, enabling comparison of safety and continual learning mechanisms while controlling for the underlying RL algorithm. The benchmarked methods, their categories, and brief descriptions are provided below and summarized in Figure \ref{fig:algorithms}; additional details appear in \ref{sec:appendix_methods}.

\noindent
\textbf{PPO} \citep{schulman2017proximal}: Proximal policy optimization. A standard on-policy RL algorithm that is used as a baseline. \\
\noindent
\textbf{CPO} \citep{achiam2017constrained}: Constrained policy optimization. A safe RL algorithm that extends TRPO with a safety constrained policy update. \\
\noindent
\textbf{PPO-Lag} \citep{ray2019benchmarking}: Lagrangian PPO. A safe RL algorithm that uses Lagrangian functions with PPO to optimize the safety constraint. \\
\noindent
\textbf{CPPO-PID} \citep{stooke2020responsive}: Constraint-controlled PPO with PID. A safe RL algorithm that extends PPO-Lag to incorporate a PID controller to adjust the Lagrangian multiplier. \\
\noindent
\textbf{PPO+EWC} \citep{nath2023modulating}: PPO with elastic weight consolidation. A continual RL algorithm that penalizes the network for modifying previously important parameters. \\
\noindent
\textbf{Experience Replay}: A continual RL algorithm inspired by CLEAR \citep{rolnick2019experience} that mixes past and present experiences during policy updates of PPO. It maintains both long- and short-term replay buffers to store information relevant to past and current tasks. \\

Based on the known properties of each baseline algorithm's class, we can hypothesize how each algorithm should behave in a non-stationary setting with safety constraints. Intuitively, the purely continual reinforcement learning methods, PPO+EWC and Experience Replay, should violate safety when they would lead to a higher or better-remembered reward. They may better remember safety constraints across tasks when they do not conflict with the task objective. The purely safe reinforcement learning methods, CPO, PPO-Lag, and CPPO-PID, should better balance safety constraints and reward while avoiding catastrophic forgetting. It is unclear whether the environment's non-stationary dynamics will affect safety. The plain reinforcement learning algorithm, PPO, should ignore safety constraints and task forgetting and be trained solely on the current task reward signal. The results and analyses of these methods are given in Section \ref{sec:results}.

\subsection{Hyperparameter Optimization}

Algorithm \ref{alg:safe-ewc} shows that each method includes multiple hyperparameters that influence performance. To ensure fair comparisons, we optimize hyperparameters on proxy tasks aligned with each method’s domain. For PPO, hyperparameters are tuned to maximize return in the nominal version of each environment. For safe RL methods, hyperparameters are tuned to maximize return while accounting for costs in nominal environments, and performance is evaluated in terms of both safety and reward. For continual RL methods, hyperparameters are tuned to maximize return under nominal dynamics and on the first new task; evaluation then measures forgetting on the nominal dynamics and performance on the new task. Hyperparameters are optimized separately for each benchmark using the Tree-Structured Parzen Estimator \citep{bergstra2011algorithms} in Optuna \citep{akiba2019optuna}. Additional details on the methods and their hyperparameters are provided in Section \ref{sec:appendix_methods}. 

\section{Results} \label{sec:results}

We trained the baselines and our proposed algorithms with their optimized hyperparameters on all three environments. To account for stochasticity, we used five random seeds per algorithm. For a single seed, a full training sequence on Damaged HalfCheetah Velocity required 4–6 hours, whereas Safe Continual World and Damaged Ant Velocity required 12–14 hours on CPU. All experiments were performed on a computer with an AMD Ryzen Threadripper 3960X 24-Core Processor and 256 GB of RAM. The following subsections report the results and discuss their implications for key questions in safe continual RL. 

\subsection{Metrics to Measure Safe Continual Reinforcement Learning}

We evaluate the extent to which each method addresses safe continual reinforcement learning using the metrics proposed in \citep{coursey2025safecrl} and \citep{wolczyk2021continual}:

\begin{itemize}
    \item \textbf{Final Task Reward}: the average undiscounted cumulative reward over the final 20 episodes for each task, measuring overall performance.
    
    \item \textbf{Total Cost}: the average total cost for each task. Calculated as follows, where $N$ is the number of times a task is visited, and costs is a vector of costs during training. This measures safety. $\frac{1}{N} \sum_{i=1}^{\text{len(costs)}} \text{costs}_i.$

    \item \textbf{Forgetting}: the difference in undiscounted cumulative reward, averaged over 20 episodes, when starting a task again from the last time that task was seen. This measures continual learning performance. If this is negative, the agent exhibited \textit{positive forward transfer} when revisiting the task. If this is positive and large, it exhibits \textit{catastrophic forgetting}. $\text{forgetting}=  r_{\text{final}} - r_{\text{immediate}}.$

    \item \textbf{Normalized Forgetting}: mean forgetting normalized by the magnitude of the reward range. The magnitude of the reward range is calculated by subtracting the initial reward from the final task reward. The rewards are averaged cumulative undiscounted rewards over 20 episodes.

    \item \textbf{Average Success Rate}: a running average task completion rate for Safe Continual World tasks during learning. A higher reward does not necessarily imply higher task completion. In the Safe Continual World environment, forgetting metrics are calculated using this instead of reward.
\end{itemize}

\subsection{Quantitative Performance on Environments}

\begin{table}[t]
\caption{Damaged HalfCheetah Velocity results. Mean $\pm$ standard deviation across 5 random seeds and all tasks.}
\label{tab:halfcheetah}
\centering
\begin{tabular}{@{}llcl@{}} 
\toprule
Algorithm   & \multicolumn{1}{c}{Final Reward ($\uparrow$)}  & \multicolumn{1}{c}{Normalized Forgetting ($\downarrow$)} & \multicolumn{1}{c}{Total Costs ($\downarrow$)} \\ \midrule
PPO         & $2078.8 \pm 559.2$               & $0.242$                      & $7368.4 \pm 2759.2$             \\
CPO         & $1536.3 \pm 120.1$                & $0.588$                     & $495.5 \pm 109.6$               \\
PPO-Lag     & $1553.5 \pm 335.1$                & $0.284$                    & $830.6 \pm 453.3$               \\
CPPO-PID    & $1487.8 \pm 330.2$                & $0.242$                       & $576.5 \pm 280.5$               \\
PPO+EWC     & $2036.6 \pm 555.9$                 & $0.242$                   & $6422.6 \pm 3476.4$             \\
Exp. Replay & $504.9 \pm 294.7$                  & $0.173$                   & $666.7 \pm 1181.9$              \\
\textbf{Safe EWC (Ours)}    & $\mathbf{1631.3 \pm 240.6}$                   & $\mathbf{0.180}$                  & $\mathbf{393.1 \pm 452.4}$ \\
\textbf{CF-EWC (Ours)} & $\mathbf{2067.1 \pm 610.0}$ & $\mathbf{0.221}$ & $\mathbf{6293.5 \pm 3416.7}$\\\bottomrule 
\end{tabular}
\end{table}

Results for Damaged HalfCheetah Velocity are reported in Table \ref{tab:halfcheetah}, averaged over three tasks and five random seeds. A safe continual RL solution should achieve high reward while minimizing forgetting and safety violations (Section \ref{sec:problem}); thus, strong performance is required across all metrics. PPO attained the highest average reward but also the highest total cost. Safe RL methods (CPO, PPO-Lag, and CPPO-PID) reduced cost relative to PPO and the continual RL baselines, but exhibited higher normalized forgetting, indicating limited continual learning. Conversely, continual RL methods (PPO+EWC and experience replay) reduced normalized forgetting relative to safe RL methods, but incurred higher total cost. Experience replay achieved comparatively low total cost largely due to sample inefficiency: because the safety constraint targets velocity, a lower reward is associated with a lower cost. Consistent with this, experience replay produced the lowest final reward, suggesting that replaying prior experience impeded learning the current task under PPO. Safe EWC achieved the second-lowest normalized forgetting while attaining the lowest cost, albeit with reduced reward, indicating a favorable safety–continual learning trade-off. In contrast, CF-EWC did not effectively prevent safety violations.

\begin{table}[t]
\caption{Damaged Ant Velocity results. Mean $\pm$ standard deviation across 5 random seeds and all tasks.}
\label{tab:ant}
\centering
\begin{tabular}{@{}llcl@{}} 
\toprule
Algorithm   & \multicolumn{1}{c}{Final Reward ($\uparrow$)} & \multicolumn{1}{c}{Normalized Forgetting ($\downarrow$)} & \multicolumn{1}{c}{Total Costs ($\downarrow$)} \\ \midrule
PPO         & $3282.1 \pm 413.8$              & $0.022$                      & $9181.0 \pm 3031.0$             \\ 
CPO         & $2708.7 \pm 117.7$                & $0.303$                     & $886.4 \pm 85.7$               \\
PPO-Lag     & $2676.4 \pm 95.0$                 & $0.001$                    & $1009.7 \pm 368.6$               \\
CPPO-PID    & $2688.8 \pm 92.7$               & $0.013$                       & $800.6 \pm 83.1$               \\ 
PPO+EWC     & $3645.3 \pm 332.4$                  & $0.008$                   & $11061.7 \pm 2273.2$             \\
\textbf{Safe EWC (Ours)}    & $\mathbf{2689.9 \pm 71.9}$                   & $\mathbf{0.051}$                  & $\mathbf{312.4 \pm 65.5}$ \\
\textbf{CF-EWC (Ours)} & $\mathbf{3739.7 \pm 266.2}$ & $\mathbf{0.006}$ & $\mathbf{11208.5 \pm 1457.2}$\\\bottomrule 
\end{tabular}
\end{table}

In Damaged HalfCheetah Velocity, the constrained control task was sufficiently simple, and all methods solved it upon first exposure, except experience replay. This was not observed in the other environments. In Damaged Ant Velocity, most methods reached near-optimal performance only toward the end of the task sequence, as reflected in Table \ref{tab:ant}. Consequently, forgetting and normalized forgetting are low; in contrast, many methods exhibit forward transfer, leveraging knowledge from the intact ant to learn damaged variants more efficiently. The primary exception is the safe RL method CPO, which shows higher forgetting; as discussed in the next subsection, CPO converges faster and transfers less across tasks.

Beyond forgetting, reward–cost trends mirror those in HalfCheetah, with Safe EWC providing the most favorable trade-off between reward and cost. CF-EWC achieves the highest average final reward but also incurs more safety violations, further suggesting that preserving ``safe” weights is insufficient to enforce safety constraints. Finally, experience replay could not be trained reliably on Ant and the Safe Continual World due to out-of-memory errors caused by larger state spaces. This highlights a limitation of replay-based approaches in realistic environments, further justifying our choice of regularization-based continual learning mechanisms.

\begin{table}[t]
\caption{Safe Continual World results. Mean $\pm$ standard deviation across 5 random seeds and all tasks.}
\label{tab:cw}
\centering
\begin{tabular}{@{}llcl@{}} 
\toprule
Algorithm   & \multicolumn{1}{c}{Success Rate ($\uparrow$)}  & \multicolumn{1}{c}{Normalized Forgetting ($\downarrow$)} & \multicolumn{1}{c}{Total Costs ($\downarrow$)} \\ \midrule
PPO         & $0.89 \pm 0.17$                & $0.440$                      & $112210.1 \pm 245724.1$             \\ 
CPO         & $0.85 \pm 0.24$               & $0.560$                     & $68817.6 \pm 92071.4$               \\
PPO-Lag     & $0.12 \pm 0.22$                & $0.025$                    & $55758.4 \pm 97358.5$               \\
CPPO-PID    & $0.01 \pm 0.02$                & $-0.618$                       & $18056.7 \pm 20715.8$               \\ 
PPO+EWC     & $0.88 \pm 0.17$                & $0.366$                   & $130252.9 \pm 207224.6$             \\
\textbf{Safe EWC (Ours)}   & $\mathbf{0.14 \pm 0.24}$                  & $\mathbf{0.070}$                  & $\mathbf{25014.1 \pm 23109.9}$ \\
\textbf{CF-EWC (Ours)} & $\mathbf{0.79 \pm 0.19}$ & $\mathbf{0.510}$ & $\mathbf{126258.7 \pm 251984.8}$\\\bottomrule 
\end{tabular}
\end{table}

In the Safe Continual World, enforcing safety constraints prevented several methods from reliably learning the tasks (Table \ref{tab:cw}). PPO, PPO+EWC, and CF-EWC, which do not explicitly penalize safety violations, achieved the highest average success rates but also the highest total costs; among them, PPO+EWC exhibited the lowest forgetting. CPO achieved a comparable success rate at approximately half the cost, but still incurred higher costs than the remaining safe methods and showed greater forgetting on task revisits. PPO-Lag, CPPO-PID, and Safe EWC achieved substantially lower success rates, reflecting a stronger emphasis on safety. This reflects an inherent difficulty of this environment in the balance between safety and task success. This trade-off will also be present in many real environments. Although CPPO-PID yielded the lowest average total cost, Safe EWC provided the most favorable trade-off among success, forgetting, and safety.  

The quantitative results highlight the need for safe, continual RL methods. Current safe RL methods tend to exhibit high levels of catastrophic forgetting, while continual RL methods that lack safety measures often compromise safety. This issue is particularly evident in the Safe Continual World environment, where safety constraints hinder task success for most safe and safe continual RL algorithms. 

\subsection{Visualizing the Tradeoff Between Metrics}

\begin{figure}[t]
    \centering

    % First subfigure
    \begin{subfigure}[b]{0.32\textwidth}
        \centering
        \includegraphics[width=\textwidth, trim=0 0 0 90, clip]{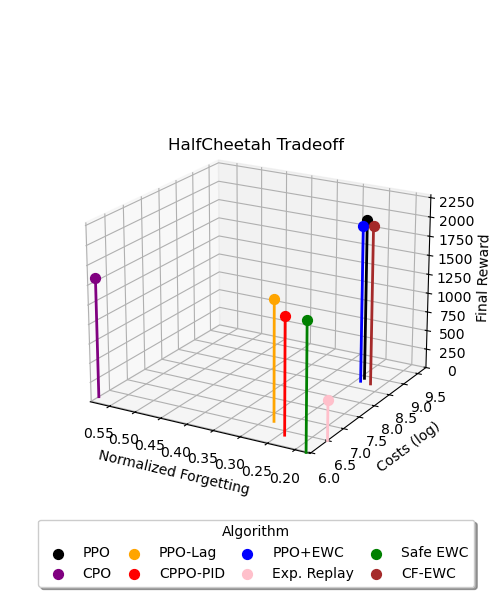} % trim=left bottom right top
        \caption{HalfCheetah}
        \label{fig:cheetah_tradeoff}
    \end{subfigure}
    \hfill
    % Second subfigure
    \begin{subfigure}[b]{0.32\textwidth}
        \centering
        \includegraphics[width=\textwidth, trim=0 0 0 90, clip]{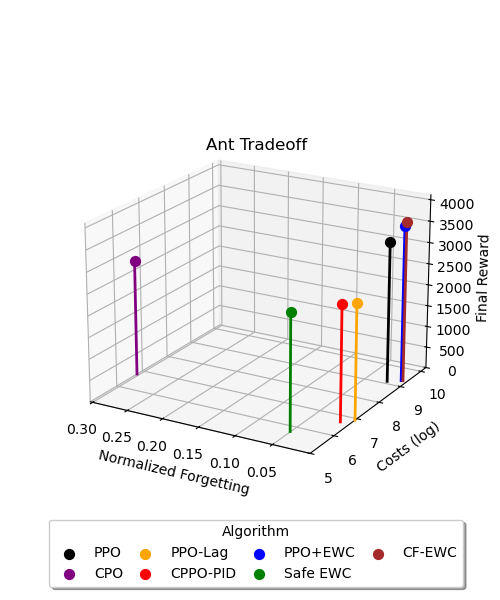}
        \caption{Ant}
        \label{fig:ant_tradeoff}
    \end{subfigure}
    \hfill
    % Third subfigure
    \begin{subfigure}[b]{0.32\textwidth}
        \centering
        \includegraphics[width=\textwidth, trim=0 0 0 90, clip]{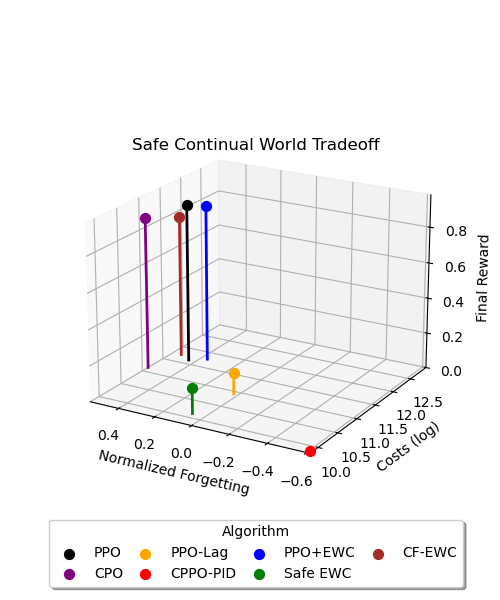}
        \caption{Safe Continual World}
        \label{fig:cw_tradeoff}
    \end{subfigure}

    \caption{Average safe continual RL metrics for the algorithms on each environment. Algorithms that better handle the tradeoff among safety, reward/success, and forgetting will be higher and closer to the front of the plot. Cost is shown on a log scale.}
    \label{fig:tradeoff}
\end{figure}

The quantitative results can be further interpreted by inspecting the trade-offs in Figure \ref{fig:tradeoff}. In HalfCheetah and Ant (Figures \ref{fig:cheetah_tradeoff} and \ref{fig:ant_tradeoff}), Safe EWC achieves a more favorable cost–forgetting trade-off than the other methods. Its average reward is slightly lower than the non-safe baselines and comparable to the safe RL methods, reflecting the inherent safety–reward trade-off: \textit{the agent must maximize distance while remaining below a velocity limit}. Similar trends appear in Safe Continual World (Figure \ref{fig:cw_tradeoff}), where the cost–forgetting trade-off is surpassed only by CPPO-PID, which attains near-zero task success. In this environment, Safe EWC achieves lower success than some methods but maintains lower cost, indicating that Safe EWC is promising for safe continual RL, though reward scaling can reduce task success. In contrast, the performance of CF-EWC consistently aligns with that of PPO and PPO+EWC. While it shows improvements in reward and reduces forgetting compared to these baselines, it does not consistently prevent safety violations. This indicates that allowing unsafe weights to be forgotten is not sufficient to ensure safe behavior, offering valuable insights for future algorithm design.

\subsection{Behavior During Lifelong Learning}

\begin{figure}[t]
    \centering
    \includegraphics[width=\linewidth]{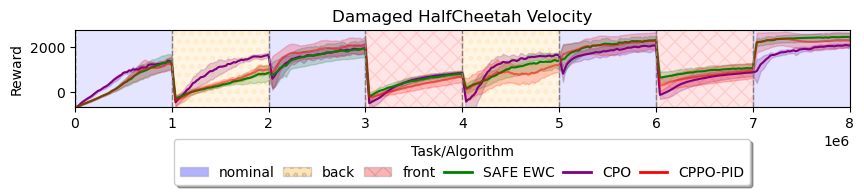}
    \includegraphics[width=\linewidth]{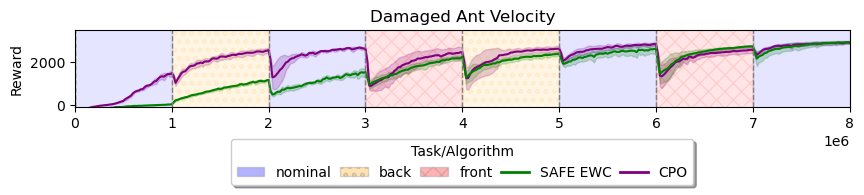}
    \vspace{-1em}
    \includegraphics[width=\linewidth]{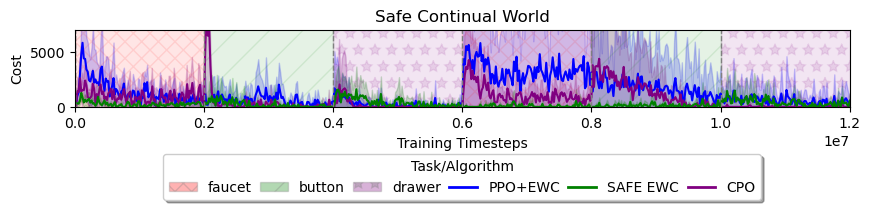}
    \caption{Behaviors during learning. \textbf{Top}: Reward behavior during training of two safe RL algorithms (CPO and CPPO-PID) versus a safe continual RL algorithm (Safe EWC) on the Damaged HalfCheetah Velocity environment. \textbf{Middle}: Reward behavior during training of a safe RL algorithm (CPO) versus a safe and continual RL algorithm (Safe EWC) on the Damaged Ant Velocity environment. \textbf{Bottom}: Cost behavior during training of a continual (PPO+EWC) versus a safe continual RL algorithm (Safe EWC).}
    \label{fig:qualitative}
\end{figure}

Additionally, we conducted a qualitative analysis to characterize how safety constraints in non-stationary environments influence training behavior during lifelong learning. Training rewards and costs are shown in Figure \ref{fig:qualitative}. For clarity, we plot a subset of algorithms; complete learning curves are provided in the Appendix (\ref{sec:appendix_isolated curves} and \ref{sec:appendix_full_curves}).

The top panel of Figure \ref{fig:qualitative} reports rewards for safe RL (CPO, CPPO-PID) and safe continual RL (Safe EWC) on Damaged HalfCheetah Velocity. Following task changes, all methods exhibit reward drops due to abrupt dynamics shifts induced by limb loss or repair. Early in training, agents typically re-enter a task with lower reward than when they last encountered it, indicating forgetting. As tasks are revisited, Safe EWC and CPPO-PID retain more knowledge and exhibit similar reward trajectories late in training, suggesting convergence to comparable policies. Although EWC’s benefits are modest in this environment (more clearly reflected in Table \ref{tab:halfcheetah}), they are most apparent upon revisiting earlier tasks. In contrast, CPO exhibits substantially greater catastrophic forgetting. Overall, the relative advantage of safe continual RL over safe RL is algorithm- and environment-dependent; here, CPPO-PID also effectively retains knowledge.

The middle panel shows reward behavior in Damaged Ant Velocity. Consistent with prior quantitative results, several methods exhibit forward transfer. In particular, Safe EWC often achieves a higher reward when revisiting a task than during its previous exposure, indicating the reuse of knowledge across tasks. CPO does not show this pattern; instead, it converges more rapidly, plausibly because its underlying RL algorithm (TRPO) is more effective than PPO in this setting. CPO also forgets more across task changes, suggesting that mitigating forgetting may be easier when performance remains far from saturation, and there is substantial capacity for improvement.

The bottom panel reports training costs for a continual RL method (PPO+EWC), a safe continual RL method (Safe EWC), and a safe RL method (CPO). Safe EWC maintains a lower average cost throughout training, particularly at task onset before CPO re-establishes safe behavior. Compared with PPO+EWC, incorporating safety into the continual learning mechanism yields consistently lower costs, and Safe EWC’s safety violations generally decrease over training. Notably, all methods exhibit increased safety violations at task onset despite unchanged safety constraints across tasks; we investigate this effect in the next subsection.

\subsection{Remembering Safety Constraints}

Next, we aim to determine whether avoiding catastrophic forgetting of rewards also prevents forgetting costs and whether safety constraint violations increase when the environment or task changes.

\begin{figure}[t]
    \centering
    \includegraphics[width=\linewidth]{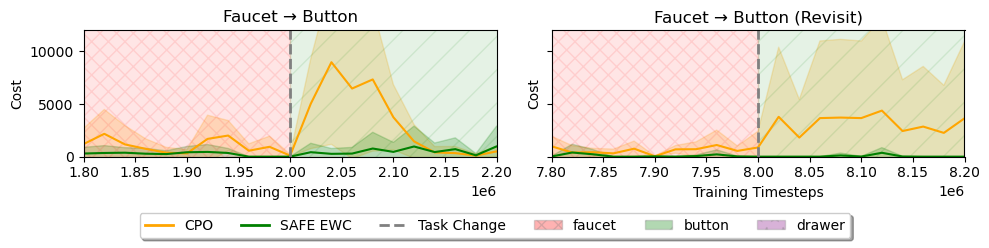}
    \caption{Cost around the moment the task changes in Safe Continual World with a safe (CPO) and safe continual (Safe EWC) algorithm. The task change triggers safety violations in CPO, which is achieving a higher reward.}
    \label{fig:task_change}
\end{figure}

To illustrate this, we refer to Figure \ref{fig:task_change}, which depicts the cost behavior during training when the task shifts in the Safe Continual World environment. When transitioning from the faucet task to the button task for the first time (as shown in the left figure), the safe reinforcement learning algorithm CPO experienced a spike in safety violations. Although the definition of safety remained consistent across tasks, the change in dynamics seemed to trigger these initial violations. Even during a second encounter with the button task later in the sequence, CPO exhibited another spike in safety violations. This suggests that CPO catastrophically forgets its adherence to safety constraints when the reward signal changes.

In contrast, the safe, continual reinforcement learning method, Safe EWC, avoided such issues and effectively maintained its safety constraints. However, Table \ref{tab:cw} shows that Safe EWC achieved a significantly lower average task success rate, with 85\% versus 14\%. This indicates that Safe EWC is likely prioritizing safety over performance. Other safe RL methods, although they displayed lower success rates, did not forget safety as drastically as CPO did.

An ideal safe continual RL algorithm should demonstrate improved cost curves while maintaining strong task performance. This should be a key goal for future methods.

\subsection{Ablation Studies}

As described in Section \ref{sec:methods}, Safe EWC enforces safety in PPO+EWC by penalizing the reward with cost. An alternative is to apply the same regularization directly to the loss. To assess this variant, we trained Safe EWC on the Damaged HalfCheetah Velocity task sequence with an additional loss penalty of $\beta \times \text{Cost}$, using $\beta=5$ (matching the reward-penalty setting). Results are reported in Table \ref{tab:safe-ewc-location}. Safety violations increased when the penalty was moved to the loss, suggesting that the cost term may be comparatively weak relative to PPO’s reward-maximization terms and the EWC parameter-change penalty. Because the (non-online) EWC penalty grows with the number of tasks, jointly balancing reward, forgetting, and cost becomes challenging. These findings indicate that, for Safe EWC, applying the cost penalty through reward shaping is better aligned with the optimization objective.

\begin{table}[t]
\caption{Safe EWC cost regularization location impact on Damaged HalfCheetah Velocity task sequence. Mean $\pm$ standard deviation across 5 random seeds and all tasks. Safe Continual World uses success rate instead of reward.}
\label{tab:safe-ewc-location}
\centering
\begin{tabular}{@{}llcl@{}} 
\toprule
Cost Penalty Location   & \multicolumn{1}{c}{Final Reward ($\uparrow$)}& \multicolumn{1}{c}{Normalized Forgetting ($\downarrow$)} & \multicolumn{1}{c}{Total Costs ($\downarrow$)} \\ \midrule
Loss Function & $2036.6 \pm 555.9$                & $0.242$                   & $6422.6 \pm 3476.4$              \\
Reward    & $1631.3 \pm 240.6$                  & $0.180$                  & $393.1 \pm 452.4$ \\\bottomrule 
\end{tabular}
\end{table}

\begin{table}[t]
\caption{Ablation building up from the core components of CF-EWC (PPO+EWC+Cost Fisher) and Safe EWC (PPO+EWC+Safe Reward). Mean $\pm$ standard deviation across 5 random seeds and all tasks.}
\label{tab:ablation}
\centering
\begin{tabular}{@{}llcl@{}} 
\toprule
Component   & \multicolumn{1}{c}{Final Reward ($\uparrow$)} &  \multicolumn{1}{c}{Normalized Forgetting ($\downarrow$)} & \multicolumn{1}{c}{Total Costs ($\downarrow$)} \\ \midrule \midrule
\multicolumn{4}{c}{Damaged HalfCheetah Velocity} \\\midrule 
PPO         & $\mathbf{2078.8 \pm 559.2}$           &$0.242$                      & $7368.4 \pm 2759.2$             \\
+EWC         & $2036.6 \pm 555.9$               & $0.242$                   & $6422.6 \pm 3476.4$             \\
+EWC+Cost Fisher     & $2067.1 \pm 610.0$  & $0.221$ & $6293.5 \pm 3416.7$\\
+EWC+Safe Reward    & $1631.3 \pm 240.6$              & $\mathbf{0.180}$                  & $\mathbf{393.1 \pm 452.4}$ \\\midrule
\multicolumn{4}{c}{Damaged Ant Velocity} \\\midrule 
PPO         & $3282.1 \pm 413.8$             & $0.022$                      & $9181.0 \pm 3031.0$             \\  
+EWC         & $3645.3 \pm 332.4$              & $0.008$                   & $11061.7 \pm 2273.2$             \\
+EWC+Cost Fisher     & $\mathbf{3739.7 \pm 266.2}$  & $\mathbf{0.006}$ & $11208.5 \pm 1457.2$\\
+EWC+Safe Reward    & $2689.9 \pm 71.9$                & $0.051$                  & $\mathbf{312.4 \pm 65.5}$ \\\midrule
\multicolumn{4}{c}{Safe Continual World} \\\midrule 
PPO         & $\mathbf{0.89 \pm 0.17}$              & $0.440$                      & $112210.1 \pm 245724.1$             \\  
+EWC         & $0.88 \pm 0.17$                & $0.366$                   & $130252.9 \pm 207224.6$             \\
+EWC+Cost Fisher     & $0.79 \pm 0.19$ & $0.510$ & $126258.7 \pm 251984.8$\\
+EWC+Safe Reward    & $0.14 \pm 0.24$               & $\mathbf{0.070}$                  & $\mathbf{25014.1 \pm 23109.9}$ \\\bottomrule 
\end{tabular}
\end{table}

In a subsequent experiment, we performed ablation on the proposed methods, Safe EWC and CF-EWC (Table \ref{tab:ablation}). We report metrics for vanilla PPO, PPO with an EWC regularizer (PPO+EWC), PPO+EWC with the Cost-Fisher modification (CF-EWC), and PPO+EWC with cost-based reward penalization (Safe EWC). Across settings, adding EWC to PPO generally reduced catastrophic forgetting while preserving comparable reward. Incorporating the Cost-Fisher modification further reduced forgetting in most cases (except Safe Continual World) by permitting forgetting of parameters associated with higher safety violations. This suggests that CF-EWC increases network plasticity, mitigating a common limitation of EWC: the capacity to learn new tasks diminished over time. However, reweighting the Fisher information in CF-EWC did not reliably improve safety, indicating that forgetting parameters correlated with unsafe behavior was insufficient to prevent future violations. Future work should examine whether safety-relevant knowledge can instead be explicitly retained.

In contrast, adding the cost-based reward penalty to PPO+EWC substantially improved safety satisfaction while maintaining low forgetting and competitive reward, supporting Safe EWC as a promising safe continual RL approach. In Safe Continual World, this improvement came at the expense of task success, highlighting that balancing safety and performance becomes more difficult in complex environments and motivating further methodological development.

\section{Discussion and Open Questions} \label{sec:discussion}

We discuss the implications of our empirical studies, open questions and future work, and limitations of this study.

\subsection{Implications}
\textbf{Current approaches do not solve the safe continual RL problem.} Our experiments indicate that existing safe RL and continual RL methods do not adequately address safe continual reinforcement learning. Sudden changes in dynamics and conflicts between reward maximization and safety constraints make the setting particularly challenging and likely require specialized methods. While our Safe EWC method effectively balances safety with catastrophic forgetting, its reward shaping mechanism reduced task success in the Safe Continual World environment, a consequence of the difficult balance between safety and performance in this environment. At the same time, our CF-EWC approach increased reward and reduced catastrophic forgetting, but did not effectively maintain safety constraints. Future approaches should further reduce catastrophic forgetting and promote positive forward and backward transfer in non-stationary environments, while simultaneously satisfying critical safety constraints.

\textbf{Future methods should explicitly retain safety constraints.} Although continual RL techniques primarily mitigate forgetting of reward, we observe that task changes can increase cost even when safety-relevant states remain unchanged (Figure \ref{fig:task_change}). Consistent with the safe continual RL formulation, future work should develop mechanisms that explicitly preserve safety constraints across tasks.

\textbf{The tension among objectives must be resolved.} Safe continual RL requires balancing task performance, continual learning, and safety, and the appropriate trade-off is application dependent. This balance may be improved by directly optimizing a safe continual RL objective, analogous to Lagrangian formulations in constrained policy optimization. Alternatively, methods may combine mechanisms from safe RL and continual RL, such as using EWC to preserve parameters important for safety in non-stationary settings and replay buffers to rehearse safety-critical states. Extending safe RL algorithms to continual settings (e.g., Lagrangian PPO under task non-stationarity) is another promising direction. 

\subsection{Open Questions and Future Work}

Our study identifies several open questions that motivate future research in safe continual reinforcement learning:

\textit{What constitutes an appropriate safe continual RL benchmark?} Selecting suitable benchmarks is nontrivial. In many standard RL environments (e.g., CartPole), adapting to new dynamics is relatively simple, making it hard to meaningfully evaluate continual learning. Conversely, combining non-stationarity with safety constraints (e.g., Safe Continual World) can make the problem prohibitively difficult. Future work should develop additional environments that complement those introduced here and provide practical, discriminative benchmarks for algorithm development.

\textit{When is continual adaptation necessary?} In HalfCheetah, a safe RL method (CPPO-PID) and other PPO-based approaches exhibited limited catastrophic forgetting. In settings with gradual dynamics shifts, adapting a safe RL method may be preferable to the additional computation required by safe continual RL methods. Future work should characterize when continual adaptation is necessary and when simpler adaptations suffice. 

\textit{Which mechanisms are effective for safe continual RL?} Because safe continual RL integrates the goals of safe RL and continual RL, progress will likely draw on mechanisms from both areas. Safe RL methods may be extended with replay, weight consolidation, or network expansion, while continual RL methods may be augmented to satisfy safety constraints. Systematically identifying and adapting effective mechanisms from each literature is a key direction.

\textit{How should methods handle hard safety constraints?} This work considers soft constraints, where violations are penalized but not strictly forbidden. Many applications require hard constraints that must never be violated (e.g., rocket engine control). Ensuring hard safety under continual learning may require approaches that strictly prevent forgetting of safety-critical behavior. Developing algorithms for this regime remains an important open problem. 

\textit{How will imperfect task detection impact safe continual reinforcement learning?} In this study, we assumed that task changes could be reliably detected, a reasonable assumption given the dramatic nature of our environment changes. However, in some real domains, the nature of the non-stationarity will be more subtle. In these domains, the core problem of safe continual RL will likely be easier, as environment changes that do not significantly impact performance will require less adaptation. At the same time, task detection will be more difficult, potentially impacting the performance of continual learning mechanisms that rely on task detection, like EWC. Future work should evaluate task detection methods and study the impact of environment changes on the tradeoff between continual adaptation and task detection.

\subsection{Limitations}

While we have made an effort to systematically explore the issue of safe continual RL, this work has limitations. Notably, we evaluated a limited set of widely adopted safe and continual RL algorithms, all of which are on-policy algorithms. This approach allowed us to make fair comparisons and examine the safe and continual learning properties of these algorithms. Furthermore, \textbf{this category of algorithms is representative of those needed to control complex, non-stationary systems in real time without prior models or the benefit of offline training}. Future studies should consider other classes of RL algorithms, such as off-policy and offline RL algorithms. Our understanding of the challenges associated with safe continual RL will improve as more algorithms are tested and developed in this area.
\section{Conclusion} \label{sec:conclusion}

Reinforcement learning methods have been successful in controlling complex systems without accurate physical models; however, most algorithms assume stationary environments and do not account for unexpected non-stationary behavior changes in the system and its environment. Moreover, agents deployed in real settings must satisfy safety requirements. Although continual RL and safe RL address non-stationarity and safety, respectively, their intersection remains relatively understudied. This paper investigates these challenges under the unified problem of safe continual reinforcement learning and highlights a clear gap in the literature.

We formalized safe continual RL by combining non-stationary and constrained Markov decision process frameworks, and introduced three benchmarks: (i) Damaged HalfCheetah Velocity, in which a cheetah is suddenly damaged or repaired under a maximum-velocity constraint; (ii) Damaged Ant Velocity, which applies the same setting to an ant robot; and (iii) Safe Continual World, in which a robotic arm performs sequential household tasks while avoiding spilling a coffee mug. We benchmark eight algorithms spanning standard RL, safe RL, continual RL, and safe continual RL, including two proposed methods. Our results show that existing methods do not solve safe continual RL: while safe RL methods can perform well in some non-stationary settings, safe continual RL methods are often necessary, and new algorithms are required.

Among the proposed approaches, Safe EWC provides a simple and competitive baseline that balances forgetting and safety, but its cost-based reward shaping can reduce task performance. CF-EWC reduces catastrophic forgetting and can improve reward, but it does not reliably maintain safety constraints. These findings motivate methods that more effectively balance reward, safety, and forgetting. We further provide qualitative analyses of lifelong training dynamics, study the problem of retaining safety constraints across task changes (including increased safety violations at task boundaries), and perform ablations showing that Safe EWC’s penalty placement is well justified and that individual components contribute meaningfully. Finally, we outline open problems and discuss limitations to guide future work in safe continual reinforcement learning. 

\section{Acknowledgements}

This material is based upon work supported by the National Science Foundation Graduate Research Fellowship Program under Grant No. 2444112. Any opinions, findings, and conclusions or recommendations expressed in this material are those of the author(s) and do not necessarily reflect the views of the National Science Foundation.

\bibliographystyle{elsarticle-num}
\bibliography{references}

\newpage
\appendix
\section{Appendix}
\subsection{Additional Methodology Information} \label{sec:appendix_methods}

In this section, we provide additional details on the algorithms used in this paper, focusing on their objectives and constraints. We use the notation presented in the original papers, which may not be consistent with the main text of this paper. For full details, we refer readers to the cited original papers.

\textbf{Proximal Policy Optimization (PPO)} 

PPO is a standard on-policy RL algorithm. It introduces a clipped objective to penalize large policy updates

\begin{equation*}
    L^{CLIP}(\theta) = \hat{\mathbb{E}}_t \left[ \min\left( r_t(\theta) \hat{A}_t, \mathrm{clip}\left( r_t(\theta), 1-\epsilon, 1+\epsilon \right) \hat{A}_t \right) \right],
\end{equation*}

where $\hat{A}_t$ are advantage estimates and $\epsilon$ is a clipping hyperparameter. Then, the objective that is maximized every step is 

\begin{equation*}
    L_t^{PPO}(\theta) = \hat{\mathbb{E}}_t \left[ L_t^{CLIP}(\theta) - c_1 L_t^{VF}(\theta) + c_2 \mathcal{S}[\pi_\theta](s_t) \right],
\end{equation*}

where $c$ are coefficients to weight loss terms, $L_t^{VF}=(V_\theta(s_t)-V_t^{\text{targ}})^2$, and $S$ is entropy to encourage exploration. For complete details, please see \cite{schulman2017proximal}.  

\textbf{Constrained Policy Optimization (CPO)}

CPO solves a constrained reinforcement learning problem. It optimizes a constrained Markov decision process with a trust region policy optimization method. Its objective is given below.

\begin{align*}
    \pi_{k+1} &= \arg\max_{\pi \in \Pi_\theta} \mathbb{E}_{s\sim d^{\pi_k}, a\sim\pi} \left[ A^{\pi_k}(s, a) \right] \\
\text{s.t.} \ &J_{C_i}(\pi_k) + \frac{1}{1-\gamma} \mathbb{E}_{s\sim d^{\pi_k}, a\sim\pi} \left[ A^{\pi_k}_{C_i}(s, a) \right] \leq d_i \quad \forall i \\
&\bar{D}_{KL}(\pi \| \pi_k) \leq \delta.
\end{align*}

In the objective above, $C_i$ is the $i$-th cost constraint, $J_{C_i}$ is the cumulative discounted cost, $A_{C_i}$ is the cost advantage, and $d_i$ is the acceptable cumulative cost for constraint $i$. This objective is consistent with typical trust region policy optimization objectives that penalize large policy changes with the additional constraint on expected cumulative cost. In practice, this objective is not explicitly maximized. Instead, CPO proposes an approximate optimization algorithm \citep{achiam2017constrained}.

\textbf{Lagrangian PPO (PPO-Lag)}

PPO-Lag solves a constrained Markov decision process by turning it into an equivalent unconstrained optimization problem using Lagrangian methods incorporated with PPO. The objective in this method can be written as

\begin{equation*}
\max_{\theta}\min_{\lambda\ge0}  L_t^{PPO}(\theta) - \lambda L_t^{PPO_C}(\theta),
\end{equation*}

where $\lambda$ is the Lagrange multiplier that is updated using gradient descent and $L_t^{PPO_C}$ is the PPO objective using cost instead of reward. See \cite{ray2019benchmarking} for more details.

\textbf{Constraint-Controlled PPO with Proportional-Integral-Derivative control (CPPO-PID)}

CPPO solves a constrained Markov decision process using Lagrangian methods on top of PPO. However, the Lagrange multiplier is updated using PID control which may improve stability and constrain violation response speed. The $\lambda$ update rule is 

\begin{align*}
    &\Delta \gets J_C - d \\
    &\partial \gets (J_C - J_{C,prev})_+ \\
    &I \gets (I + \Delta)_+ \\
    &\lambda \gets (K_P\Delta + K_II + K_D\partial)_+, 
\end{align*}

where $K_P, K_I, K_D$ are PID parameters and $()_+$ indicates positive projection. For full details on how this update rule is incorporated with Lagrangian PPO, please see \cite{stooke2020responsive}.

\begin{algorithm}[H]
\caption{Cost Fisher Elastic Weight Consolidation (CF-EWC)}
\label{alg:cf-ewc}
\begin{algorithmic}[1]
\Require Tasks $\{\mathcal{T}_1,\dots,\mathcal{T}_K\}$
\Require PPO params $(\gamma,\lambda_{\text{GAE}},\epsilon,c_v,c_e)$
\Require Learning rates $(\alpha_\theta,\alpha_\phi)$, EWC coeff.\ $\lambda$
\Require Task length $T_{\text{task}}$, update interval $T_{\text{update}}$, Fisher samples $N$

\State Initialize $\pi_\theta$, $V_\phi$, $\mathcal{M}\leftarrow\emptyset$
    \Comment{Policy, value function, and EWC memory}

\For{$k=1$ to $K$}
    \State Initialize environment $\mathcal{T}_k$
        \Comment{Start new task}
    \State $\mathcal{D}\leftarrow\emptyset$
        \Comment{Clear rollout buffer}

    \For{$t=1$ to $T_{\text{task}}$}
        \State $a_t \sim \pi_\theta(\cdot\mid s_t)$
            \Comment{Sample action}
        \State $(s_{t+1}, r_t, C_t)\sim\mathcal{T}_k(s_t,a_t)$
            \Comment{Environment transition}
        \State $\tilde r_t \leftarrow r_t - \beta C_t$
            \Comment{Cost-shaped reward (Safe EWC)}
        \State $\mathcal{D}\leftarrow\mathcal{D}\cup(s_t,a_t,\tilde r_t,s_{t+1})$
            \Comment{Store transition}

        \If{$t \bmod T_{\text{update}} = 0$}

            \State \textbf{PPO return and advantage computation}
            \vspace{0.2cm}
            
            \State $\delta_t \leftarrow \tilde r_t + \gamma V_\phi(s_{t+1}) - V_\phi(s_t)$
                \Comment{TD residual}
            \State $\hat A_t \leftarrow \sum_l (\gamma\lambda_{\text{GAE}})^l\delta_{t+l}$
                \Comment{Generalized advantage estimation (GAE)}
            \State $G_t \leftarrow \sum_l \gamma^l\tilde r_{t+l}$
                \Comment{Discounted returns}
            \vspace{0.2cm}
            \State \textbf{PPO policy and value losses}
            \vspace{0.2cm}
            \State $r_t(\theta)\leftarrow
            \pi_\theta(a_t|s_t)/\pi_{\theta_{\text{old}}}(a_t|s_t)$
                \Comment{Importance ratio}
            \State $\mathcal{L}_{\text{PPO}}\leftarrow
            -\mathbb{E}[\min(r_t\hat A_t,\text{clip}(r_t,1\pm\epsilon)\hat A_t)]$
                \Comment{Clipped surrogate objective}
            \State $\mathcal{L}_{\text{value}}\leftarrow
            \mathbb{E}[(V_\phi(s_t)-G_t)^2]$
                \Comment{Value regression}
            \State $\mathcal{L}_{\text{entropy}}\leftarrow
            -\mathbb{E}[\mathcal{H}(\pi_\theta(\cdot|s_t))]$
                \Comment{Entropy regularization}

            \vspace{0.2cm}
            \State \textbf{Elastic Weight Consolidation (EWC)}
            \vspace{0.2cm}
            
            \State $\mathcal{L}_{\text{EWC}}\leftarrow
            \sum_{(\theta^*_j,F_j)\in\mathcal{M}}
            \sum_i \frac{\lambda}{2}F_{j,i}(\theta_i-\theta^*_{j,i})^2$
                \Comment{Penalty for forgetting previous tasks}

            \vspace{0.2cm}
            \State \textbf{Parameter updates}
            \vspace{0.2cm}
            
            \State $\mathcal{L}_\theta\leftarrow
            \mathcal{L}_{\text{PPO}}+c_e\mathcal{L}_{\text{entropy}}+\mathcal{L}_{\text{EWC}}$
                \Comment{Total actor loss}
            \State $\mathcal{L}_\phi\leftarrow c_v\mathcal{L}_{\text{value}}$
                \Comment{Total critic loss}
            \State $\theta\leftarrow\theta-\alpha_\theta\nabla_\theta\mathcal{L}_\theta$
                \Comment{Update policy}
            \State $\phi\leftarrow\phi-\alpha_\phi\nabla_\phi\mathcal{L}_\phi$
                \Comment{Update value function}

            \State $\theta_{\text{old}}\leftarrow\theta,\ \mathcal{D}\leftarrow\emptyset$
                \Comment{Prepare next rollout}
        \EndIf
    \EndFor
    
    \State $\theta^*_k\leftarrow\theta$
        \Comment{Save task-optimal parameters}
    \State Sample $N$ transitions $\{(s_n, a_n, c_n)\}_{n=1}^N$ from $\mathcal{T}_k$ using $\pi_{\theta^*_k}$
        \Comment{Fisher rollout}
    \State $F_{k,i}\leftarrow
    \frac{1}{N}\sum_n\frac{1}{1+c_n}(\partial_{\theta_i}\log\pi_{\theta^*_k}(a_n|s_n))^2$
        \Comment{Estimate Cost Fisher information}
    \State $\mathcal{M}\leftarrow\mathcal{M}\cup(\theta^*_k,F_k)$
        \Comment{Store task importance}
\EndFor
\end{algorithmic}
\end{algorithm}

\textbf{PPO with Elastic Weight Consolidation (PPO+EWC)}

PPO+EWC incorporates elastic weight consolidation (EWC) \citep{kirkpatrick2017overcoming} into PPO. EWC reduces forgetting in continual learning by penalizing the network for making large changes to parameters that were important on previous tasks. The loss regularization is shown below

\begin{equation*}
    \mathcal{L}(\theta) = \mathcal{L}_B(\theta) + \sum_i\frac{\lambda}{2}F_i(\theta_i-\theta^*_{A,i})^2,
\end{equation*}

where $\mathcal{L}_B$ is the loss on the current task, $F$ is a fisher information matrix measuring the ``importance'' of each parameter, $\lambda$ is a hyperparameter to balance current and previous tasks, and $\theta_A$ are parameters for a previous task. In practice, $F$ is approximated.

To combine this with PPO, we add this regularization to PPO's loss. We assume known task boundaries and accumulate this loss over each prior task. This method is commonly used as a benchmark in continual RL \citep{nath2023modulating}. A variant of this method that may be more suitable for lifelong learning, Online EWC \citep{schwarz2018progress}, has been developed. This keeps updating a single $F$, rather than maintaining one for each task. However, since our benchmarked task sequences are short (at most 15 tasks), we did not encounter scalability issues with EWC. 

\textbf{Experience Replay}

Continual RL methods commonly use replay buffers to recall past experiences.  To develop a representative algorithm from this category, we take inspiration from CLEAR \citep{rolnick2019experience}. We maintain a buffer of the past 5 million experiences. At every PPO policy update, the agent uses a $50/50$ split of sampled and current experiences, as recommended by \cite{rolnick2019experience}. We do not include the behavior cloning loss terms from CLEAR, as those would further reduce the already-poor sample efficiency. Notably, CLEAR also uses IMPALA, a distributed off-policy RL algorithm, instead of PPO. We use PPO for consistency with the other benchmarked algorithms, but future work should incorporate experience replay with off-policy algorithms to address the sample efficiency concerns.

\subsubsection{Hyperparameters}

\begin{table}[H]
\centering
\caption{Selected HalfCheetah hyperparameters. * indicates they were chosen through hyperparameter optimization. Others were kept consistent with standard implementations. Experience Replay used PPO's hyperparameters, as there were no new parameters associated with the replay buffer. Safe EWC used PPO+EWC's hyperparameters with the cost weight. CF-EWC used PPO+EWC's hyperparameters, as there were no new hyperparameters associated with this algorithm.}
\label{tab:hyperparameters_cheetah}
\begin{tabular}{@{}llllllllll@{}}
\\
\toprule
Algorithm         & Neurons & LR & Batch Size & Epochs & Target KL & EWC $\lambda$ & Step Frac. & Lag. LR & Cost $\beta$ \\ \midrule
PPO               & 32*     & 3e-4          & 256*       & 44*                 & 0.02      & -             & -                 & -           & -            \\
CPO               & 128*    & 1e-3          & 128*       & 10                  & 0.01      & -             & 0.653*            & -           & -            \\
PPO-Lag           & 32*     & 3e-4          & 128*       & 40                  & 0.02      & -             & -                 & 0.034*      & -            \\
CPPO-PPID         & 32*     & 3e-4          & 128*       & 40                  & 0.02      & -             & -                 & -           & -            \\
PPO+EWC           & 32*     & 3e-4          & 64*        & 36*                 & 0.02      & 12.926*       & -                 & -           & -            \\
Exp. Replay & 32      & 3e-4          & 256        & 44                  & 0.02      & -             & -                 & -           & -            \\
Safe EWC          & 32      & 3e-4          & 64         & 36                  & 0.02      & 12.926        & -                 & -           & 5            \\
CF-EWC & 32     & 3e-4          & 64        & 36                 & 0.02      & 12.926       & -                 & -           & - \\\bottomrule \\
\end{tabular}
\end{table}

\begin{table}[H]
\centering
\caption{Selected Damaged Ant Velocity hyperparameters. * indicates they were chosen through hyperparameter optimization. Others were kept consistent with standard implementations. Safe EWC used PPO+EWC's hyperparameters with the cost weight. CF-EWC used PPO+EWC's hyperparameters, as there were no new hyperparameters associated with this algorithm.}
\label{tab:hyperparameters_ant}
\begin{tabular}{@{}llllllllll@{}}
\\
\toprule
Algorithm         & Neurons & LR & Batch Size & Epochs & Target KL & EWC $\lambda$ & Step Frac. & Lag. LR & Cost $\beta$ \\ \midrule
PPO               & 32*     & 3e-4          & 256*       & 43*                 & 0.02      & -             & -                 & -           & -            \\
CPO               & 64*    & 1e-3          & 128*       & 10                  & 0.01      & -             & 0.718*            & -           & -            \\
PPO-Lag           & 64*     & 3e-4          & 256*       & 40                  & 0.02      & -             & -                 & 0.025*      & -            \\
CPPO-PPID         & 128*     & 3e-4          & 256*       & 40                  & 0.02      & -             & -                 & -           & -            \\
PPO+EWC           & 64*     & 3e-4          & 64*        & 35*                 & 0.02      & 11.24*       & -                 & -           & -            \\
Safe EWC          & 64     & 3e-4          & 64        & 35                 & 0.02      & 11.24       & -                 & -           & 5              \\
CF-EWC & 64     & 3e-4          & 64        & 35                 & 0.02      & 11.24       & -                 & -           & -  \\\bottomrule \\
\end{tabular}
\end{table}

\begin{table}[H]
\centering
\caption{Selected Safe Continual World hyperparameters. * indicates they were chosen through hyperparameter optimization. Others were kept consistent with standard implementations. Safe EWC used PPO+EWC's hyperparameters with the cost weight. CF-EWC used PPO+EWC's hyperparameters, as there were no new hyperparameters associated with this algorithm.}
\label{tab:hyperparameters_cw}
\begin{tabular}{@{}llllllllll@{}}
\\
\toprule
Algorithm         & Neurons & LR & Batch Size & Epochs & Target KL & EWC $\lambda$ & Step Frac. & Lag. LR & Cost $\beta$ \\ \midrule
PPO               & 32*     & 3e-4          & 128*       & 38*                 & 0.02      & -             & -                 & -           & -            \\
CPO               & 64*    & 1e-3          & 128*       & 10                  & 0.01      & -             & 0.841*            & -           & -            \\
PPO-Lag           & 128*     & 3e-4          & 256*       & 40                  & 0.02      & -             & -                 & 0.032*      & -            \\
CPPO-PPID         & 32*     & 3e-4          & 64*       & 40                  & 0.02      & -             & -                 & -           & -            \\
PPO+EWC           & 128*     & 3e-4          & 64*        & 37*                 & 0.02      & 23.125       & -                 & -           & -            \\
Safe EWC          & 128     & 3e-4          & 64        & 37                 & 0.02      & 23.125       & -                 & -           & 5            \\
CF-EWC & 128     & 3e-4          & 64        & 37                 & 0.02      & 23.125       & -                 & -           & -            \\\bottomrule \\
\end{tabular}
\end{table}

Hyperparameters were chosen through a proxy task in a relevant variation of the environment. For PPO, this was the first task in the task sequence, and performance was measured as the final average reward. For the safe RL algorithms, this was the first task in the task sequence with the safety constraint being considered, and performance was measured as the final average reward penalized by the final average cost. For the continual RL algorithms, the proxy task consisted of the first two tasks in the respective task sequence, and performance was measured by averaging the final reward on the second task with the final reward when revisiting the first task. All hyperparameters were evaluated on 5 episodes of their evaluation environments. On the MuJoCo Damaged Velocity tasks, algorithms were trained for 500,000 timesteps per task, and on the Safe Continual World environment, algorithms were trained for 1,000,000 timesteps per task. Hyperparameters were optimized using 50 trials of the Tree-Structured Parzen estimator algorithm.

We chose these limited-interaction proxy tasks for two reasons. First, it was much more computationally feasible. A single training run on the Safe Continual World took almost a full day to finish. With 50 trials of 8 algorithms, it was infeasible to optimize hyperparameters on the full task given our budget. Second, we believe this represents the requirements of real lifelong RL will be done. In the real world, we will not know all possible environmental conditions or tasks the agent will encounter. We need to develop techniques that can adapt to these unexpected scenarios. The soft safety constraint cost limit for the safe RL algorithms was set to 25 for both environments. This value was used for the (non-continuous) HalfCheetah Velocity task in \cite{ji2023safety}. Since we tolerate some tilting of the mug in Safe Continual World (25 degrees should be fine), we kept this value consistent. In practice, this tolerance will need to be set by a domain expert or operator.

\setlength{\tabcolsep}{4pt}
\begin{table}[H]
\centering
\small
\caption{Damaged HalfCheetah Velocity results broken down by task across 5 random seeds.}
\label{tab:cheetah_task}
\begin{tabular}{@{}lccc@{}}
\toprule
Algorithm & Nominal & Front & Back \\
\midrule
\multicolumn{4}{@{}c}{Rewards} \\
\midrule
PPO & $3891.53 \pm 582.53$ & $1256.27 \pm 283.73$ & $1088.74 \pm 811.28$ \\
CPO & $2075.92 \pm 97.09$ & $875.85 \pm 170.29$ & $1657.17 \pm 92.78$ \\
PPO-Lag & $2328.29 \pm 343.64$ & $1038.47 \pm 292.45$ & $1293.82 \pm 369.24$ \\
CPPO-PID & $2311.46 \pm 340.71$ & $952.68 \pm 340.51$ & $1199.30 \pm 309.34$ \\
PPO+EWC & $3598.60 \pm 1149.40$ & $1110.57 \pm 159.67$ & $1400.64 \pm 358.69$ \\
Exp. Replay & $883.42 \pm 615.77$ & $420.81 \pm 47.21$ & $210.46 \pm 221.12$ \\
Safe EWC & $2449.72 \pm 178.80$ & $1065.29 \pm 200.83$ & $1378.86 \pm 342.28$ \\
CF-EWC & $3460.39 \pm 1039.57$ & $1152.77 \pm 208.90$ & $1588.19 \pm 581.62$ \\
\midrule
\multicolumn{4}{@{}c}{Reward Forgetting} \\
\midrule
PPO & $590.63 \pm 274.95$ & $265.65 \pm 89.17$ & $1124.30 \pm 745.13$ \\
CPO & $813.07 \pm 271.68$ & $976.40 \pm 162.72$ & $2077.69 \pm 463.75$ \\
PPO-Lag & $283.84 \pm 220.18$ & $455.21 \pm 282.71$ & $1138.59 \pm 262.72$ \\
CPPO-PID & $273.19 \pm 237.89$ & $407.43 \pm 185.74$ & $870.66 \pm 479.36$ \\
PPO+EWC & $694.21 \pm 396.63$ & $349.79 \pm 131.32$ & $902.99 \pm 475.02$ \\
Exp. Replay & $219.89 \pm 118.53$ & $183.54 \pm 52.96$ & $195.20 \pm 63.78$ \\
Safe EWC & $363.18 \pm 203.76$ & $165.10 \pm 185.93$ & $703.44 \pm 412.76$ \\
CF-EWC & $709.06 \pm 293.04$ & $190.60 \pm 120.61$ & $906.40 \pm 391.82$ \\
\midrule
\multicolumn{4}{@{}c}{Total Costs} \\
\midrule
PPO & $20006.78 \pm 7114.96$ & $902.52 \pm 501.56$ & $1195.76 \pm 660.97$ \\
CPO & $832.28 \pm 89.43$ & $39.03 \pm 38.71$ & $615.32 \pm 200.59$ \\
PPO-Lag & $1611.43 \pm 591.40$ & $76.93 \pm 110.17$ & $803.41 \pm 658.43$ \\
CPPO-PID & $1074.55 \pm 329.11$ & $118.73 \pm 148.02$ & $536.22 \pm 364.39$ \\
PPO+EWC & $17594.28 \pm 9204.50$ & $656.25 \pm 679.35$ & $1017.40 \pm 545.22$ \\
Exp. Replay & $1871.40 \pm 3453.78$ & $103.69 \pm 70.97$ & $25.08 \pm 20.92$ \\
Safe EWC & $433.28 \pm 137.70$ & $358.72 \pm 612.60$ & $387.35 \pm 606.91$ \\
CF-EWC & $17037.16 \pm 8953.20$ & $416.71 \pm 198.48$ & $1426.70 \pm 1098.33$ \\
\bottomrule
\end{tabular}
\end{table}

\subsection{Results By Task} \label{sec:appendix_tables}

In this section, we break down the results by task. This shows some behaviors better, such as the positive forward transfer (negative forgetting) of many algorithms on the Damaged Ant Velocity environment (see Table \ref{tab:ant_task}). These results also inform us which tasks were difficult for each algorithm, allowing us to focus future research on the challenges of each environment.

\setlength{\tabcolsep}{4pt}
\begin{table}[H]
\centering
\small
\caption{Damaged Ant Velocity results broken down by task across 5 random seeds.}
\label{tab:ant_task}
\begin{tabular}{@{}lccc@{}}
\toprule
Algorithm & Nominal & Front & Back \\
\midrule
\multicolumn{4}{@{}c}{Rewards} \\
\midrule
PPO & $4076.32 \pm 679.24$ & $2896.44 \pm 227.15$ & $2873.43 \pm 335.06$ \\
CPO & $2901.45 \pm 59.00$ & $2590.21 \pm 152.32$ & $2634.49 \pm 141.76$ \\
PPO-Lag & $2969.05 \pm 84.09$ & $2687.86 \pm 142.78$ & $2372.26 \pm 58.08$ \\
CPPO-PID & $2952.81 \pm 76.19$ & $2730.06 \pm 65.57$ & $2383.55 \pm 136.47$ \\
PPO+EWC & $4851.58 \pm 349.04$ & $3192.73 \pm 318.15$ & $2891.70 \pm 329.99$ \\
Safe EWC & $2932.87 \pm 75.92$ & $2751.92 \pm 41.90$ & $2384.82 \pm 97.87$ \\
CF-EWC & $4949.44 \pm 491.74$ & $3169.52 \pm 186.40$ & $3100.21 \pm 120.54$ \\
\midrule
\multicolumn{4}{@{}c}{Reward Forgetting} \\
\midrule
PPO & $-211.36 \pm 123.71$ & $602.20 \pm 310.25$ & $-148.26 \pm 444.89$ \\
CPO & $357.80 \pm 175.35$ & $1074.53 \pm 428.15$ & $1310.19 \pm 538.25$ \\
PPO-Lag & $-221.60 \pm 169.95$ & $537.47 \pm 379.97$ & $-310.48 \pm 222.81$ \\
CPPO-PID & $-200.12 \pm 29.52$ & $374.80 \pm 302.02$ & $-53.81 \pm 316.14$ \\
PPO+EWC & $-327.56 \pm 212.11$ & $677.37 \pm 388.72$ & $-249.85 \pm 295.31$ \\
Safe EWC & $-198.40 \pm 115.66$ & $782.60 \pm 438.30$ & $-114.73 \pm 284.81$ \\
CF-EWC & $-311.30 \pm 146.64$ & $594.84 \pm 475.21$ & $-207.49 \pm 218.26$ \\
\midrule
\multicolumn{4}{@{}c}{Total Costs} \\
\midrule
PPO & $14360.01 \pm 4491.68$ & $7807.25 \pm 2395.19$ & $5375.82 \pm 2206.19$ \\
CPO & $873.60 \pm 78.37$ & $871.63 \pm 34.05$ & $913.92 \pm 144.68$ \\
PPO-Lag & $1018.68 \pm 235.86$ & $1072.37 \pm 520.29$ & $938.10 \pm 349.62$ \\
CPPO-PID & $841.56 \pm 39.28$ & $970.43 \pm 135.86$ & $589.68 \pm 74.08$ \\
PPO+EWC & $17905.56 \pm 1082.56$ & $9976.02 \pm 3907.85$ & $5303.56 \pm 1829.13$ \\
Safe EWC & $362.81 \pm 21.45$ & $315.63 \pm 89.78$ & $258.84 \pm 85.15$ \\
CF-EWC & $17737.34 \pm 2147.46$ & $9659.46 \pm 1285.12$ & $6228.70 \pm 939.01$ \\
\bottomrule
\end{tabular}
\end{table}

\setlength{\tabcolsep}{4pt}
\begin{table}[H]
\centering
\small
\caption{Safe Continual World results broken down by task across 5 random seeds.}
\label{tab:cw_task}
\begin{tabular}{@{}lccc@{}}
\toprule
Algorithm & Faucet & Button & Drawer \\
\midrule
\multicolumn{4}{@{}c}{Success} \\
\midrule
PPO & $0.90 \pm 0.10$ & $1.00 \pm 0.00$ & $0.77 \pm 0.22$ \\
CPO & $0.65 \pm 0.29$ & $1.00 \pm 0.00$ & $0.90 \pm 0.17$ \\
PPO-Lag & $0.04 \pm 0.07$ & $0.28 \pm 0.32$ & $0.04 \pm 0.08$ \\
CPPO-PID & $0.00 \pm 0.00$ & $0.03 \pm 0.03$ & $0.00 \pm 0.00$ \\
PPO+EWC & $0.79 \pm 0.21$ & $1.00 \pm 0.00$ & $0.85 \pm 0.13$ \\
Safe EWC & $0.03 \pm 0.04$ & $0.39 \pm 0.29$ & $0.00 \pm 0.00$ \\
CF-EWC & $0.67 \pm 0.19$ & $0.99 \pm 0.01$ & $0.71 \pm 0.10$ \\
\midrule
\multicolumn{4}{@{}c}{Success Forgetting} \\
\midrule
PPO & $0.76 \pm 0.21$ & $0.01 \pm 0.02$ & $0.40 \pm 0.33$ \\
CPO & $0.37 \pm 0.34$ & $0.24 \pm 0.38$ & $0.81 \pm 0.15$ \\
PPO-Lag & $0.02 \pm 0.03$ & $0.02 \pm 0.11$ & $-0.03 \pm 0.05$ \\
CPPO-PID & $-0.01 \pm 0.03$ & $-0.01 \pm 0.05$ & $0.00 \pm 0.00$ \\
PPO+EWC & $0.70 \pm 0.18$ & $0.01 \pm 0.02$ & $0.25 \pm 0.31$ \\
Safe EWC & $-0.01 \pm 0.04$ & $0.07 \pm 0.13$ & $-0.03 \pm 0.07$ \\
CF-EWC & $0.72 \pm 0.29$ & $0.01 \pm 0.02$ & $0.48 \pm 0.18$ \\
\midrule
\multicolumn{4}{@{}c}{Total Costs} \\
\midrule
PPO & $289830.68 \pm 362416.12$ & $4295.64 \pm 6958.20$ & $42503.87 \pm 41154.58$ \\
CPO & $106108.08 \pm 39016.20$ & $8388.14 \pm 14744.49$ & $91956.57 \pm 134588.45$ \\
PPO-Lag & $30670.84 \pm 12497.80$ & $121709.39 \pm 146814.47$ & $14895.11 \pm 8754.31$ \\
CPPO-PID & $23787.35 \pm 18894.64$ & $4843.23 \pm 6352.29$ & $25539.48 \pm 25032.90$ \\
PPO+EWC & $236857.96 \pm 264564.40$ & $39556.89 \pm 77616.03$ & $114343.79 \pm 181560.02$ \\
Safe EWC & $16749.54 \pm 4495.90$ & $42171.93 \pm 30738.53$ & $16120.78 \pm 13976.51$ \\
CF-EWC & $215464.75 \pm 288290.20$ & $6587.91 \pm 7541.14$ & $156723.51 \pm 290024.05$ \\
\bottomrule
\end{tabular}
\end{table}

\subsection{Isolated Training Curves} \label{sec:appendix_isolated curves}

In this section, we show the training curves of each algorithm on each environment, giving us a complete view of the behavior during lifelong learning.

\begin{figure}[H]
    \centering
    \captionsetup{justification=centering}
    \subcaptionbox{Reward for PPO on Damaged HalfCheetah Velocity.\label{fig:ppo_reward}}{
        \includegraphics[width=0.48\linewidth]{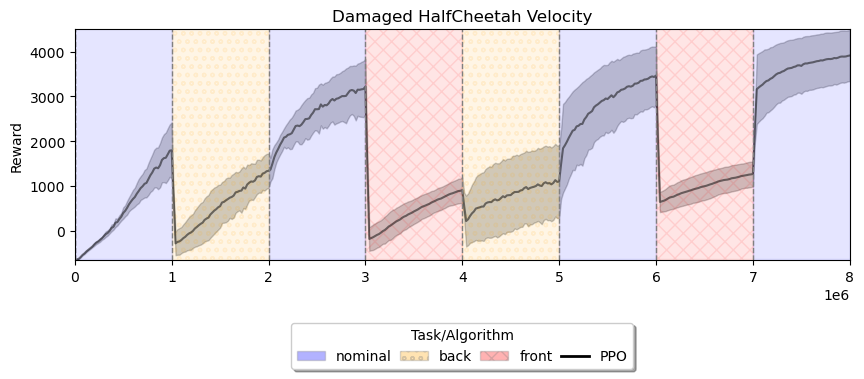}}
    \subcaptionbox{Reward for CPO on Damaged HalfCheetah Velocity.\label{fig:cpo_reward}}{
        \includegraphics[width=0.48\linewidth]{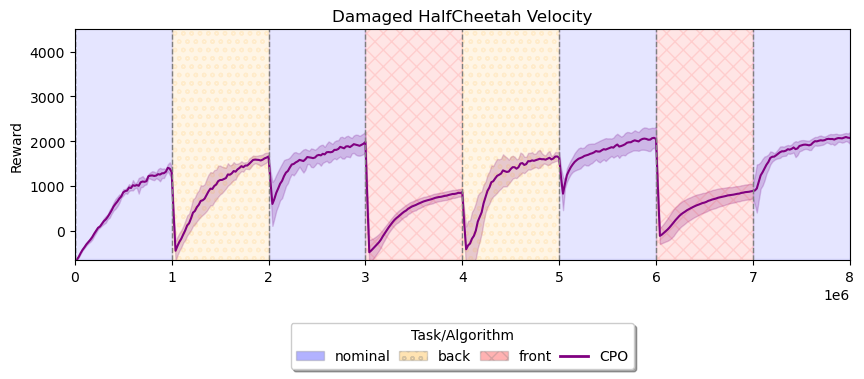}}

    \vspace{0.5cm}

    \subcaptionbox{Reward for PPO-Lag on Damaged HalfCheetah Velocity.\label{fig:ppo_lag_reward}}{
        \includegraphics[width=0.48\linewidth]{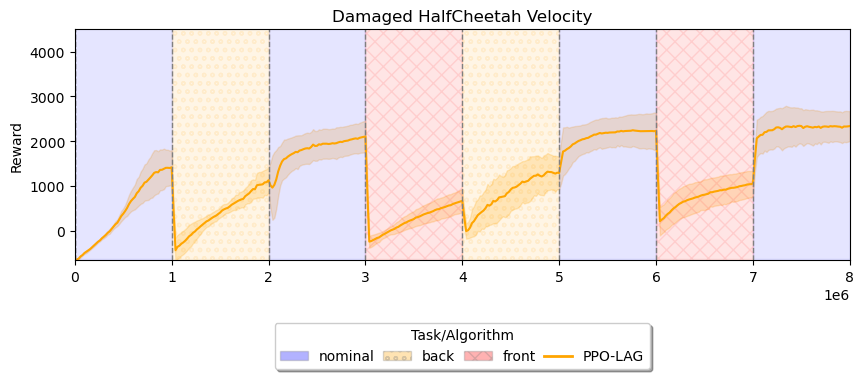}}
    \subcaptionbox{Reward for CPPO-PID on Damaged HalfCheetah Velocity.\label{fig:cppo_pid_reward}}{
        \includegraphics[width=0.48\linewidth]{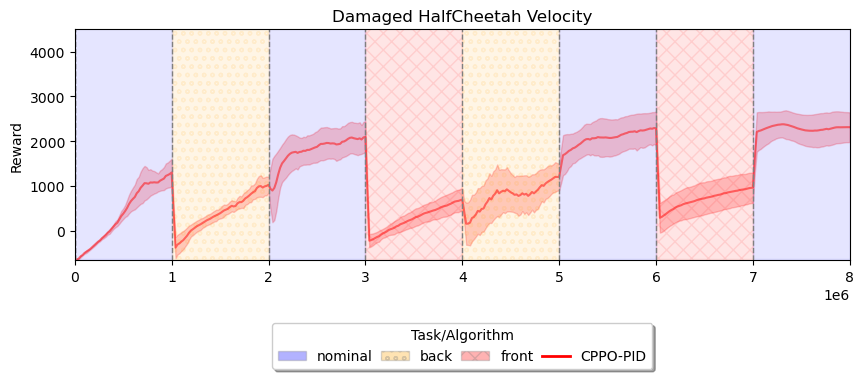}}

    \vspace{0.5cm}

    \subcaptionbox{Reward for PPO+EWC on Damaged HalfCheetah Velocity.\label{fig:ppo_ewc_reward}}{
        \includegraphics[width=0.48\linewidth]{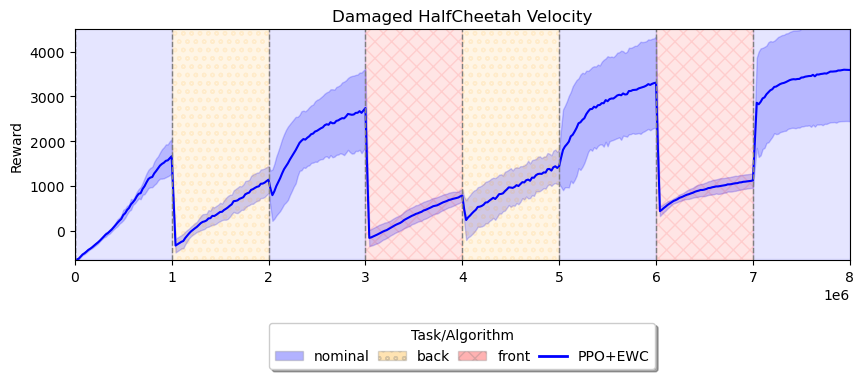}}
    \subcaptionbox{Reward for Experience Replay on Damaged HalfCheetah Velocity.\label{fig:exp_replay_reward}}{
        \includegraphics[width=0.48\linewidth]{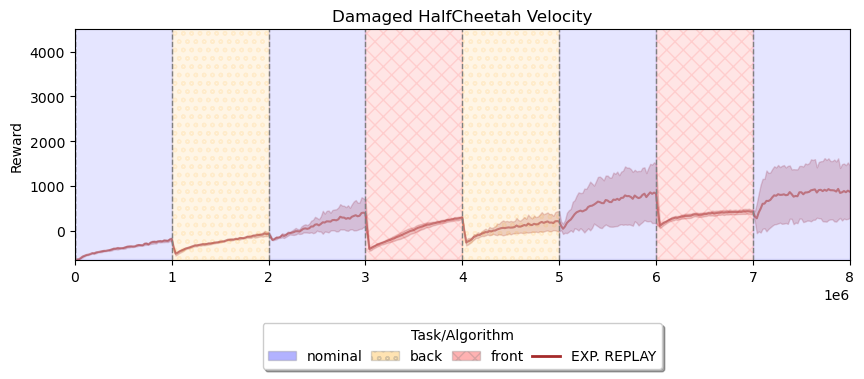}}

    \vspace{0.5cm}

    \subcaptionbox{Reward for Safe EWC on Damaged HalfCheetah Velocity.\label{fig:safe_ewc_reward}}{
        \includegraphics[width=0.48\linewidth]{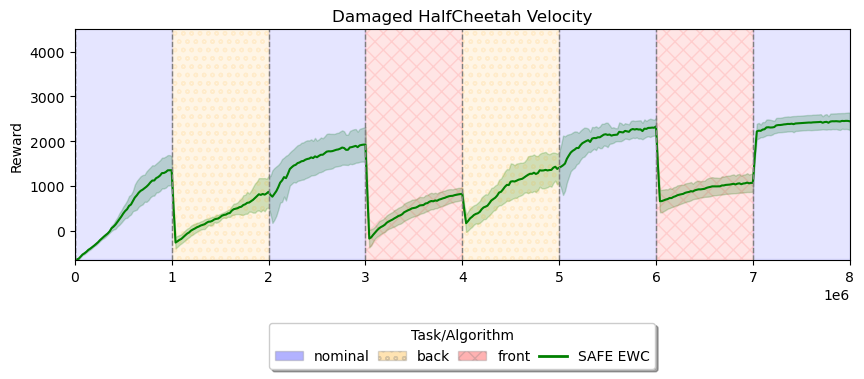}}
    \subcaptionbox{Reward for CF-EWC on Damaged HalfCheetah Velocity.\label{fig:cf_ewc_reward}}{
        \includegraphics[width=0.48\linewidth]{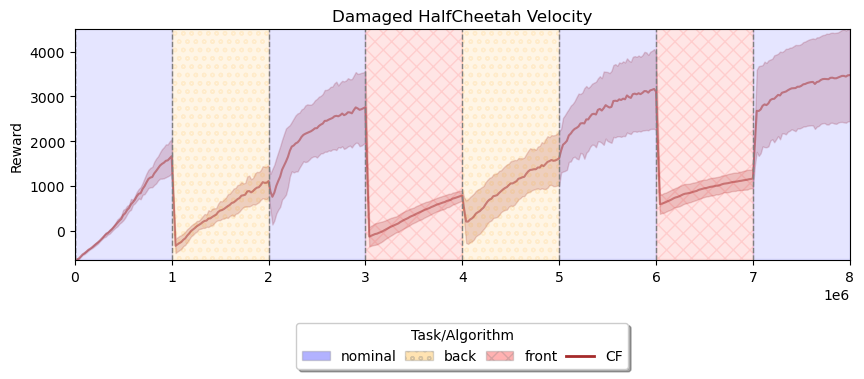}}

    \caption{Reward curves for different algorithms across 5 random seeds on Damaged HalfCheetah Velocity task.}
    \label{fig:all_rewards}
\end{figure}

\begin{figure}[H]
    \centering
    \captionsetup{justification=centering}
    \subcaptionbox{Cost for PPO on Damaged HalfCheetah Velocity.\label{fig:ppo_cost}}{
        \includegraphics[width=0.48\linewidth]{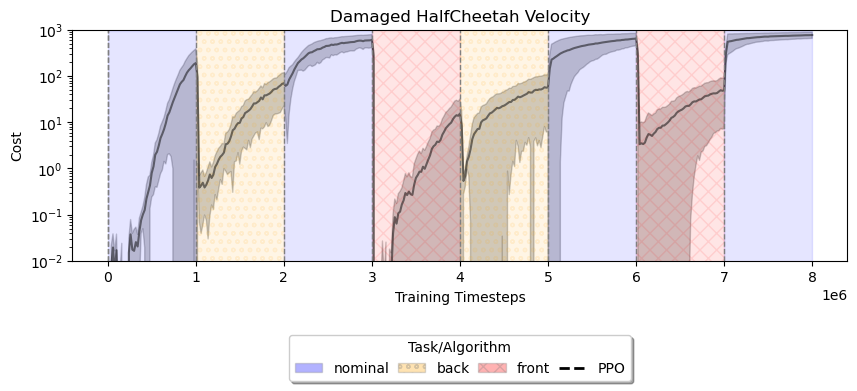}}
    \subcaptionbox{Cost for CPO on Damaged HalfCheetah Velocity.\label{fig:cpo_cost}}{
        \includegraphics[width=0.48\linewidth]{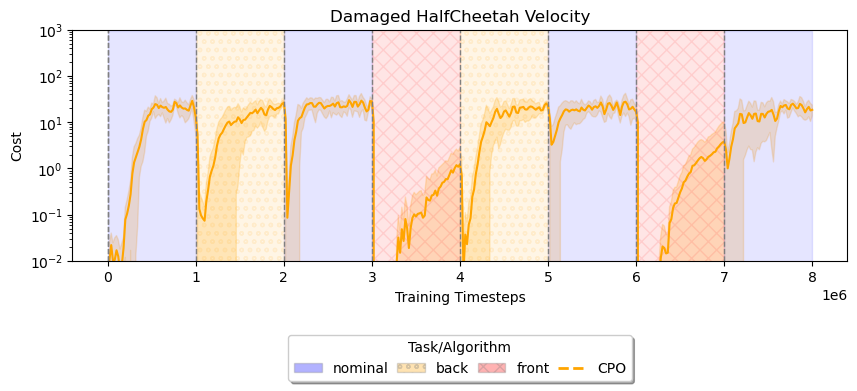}}

    \vspace{0.5cm}

    \subcaptionbox{Cost for PPO-Lag on Damaged HalfCheetah Velocity.\label{fig:ppo_lag_cost}}{
        \includegraphics[width=0.48\linewidth]{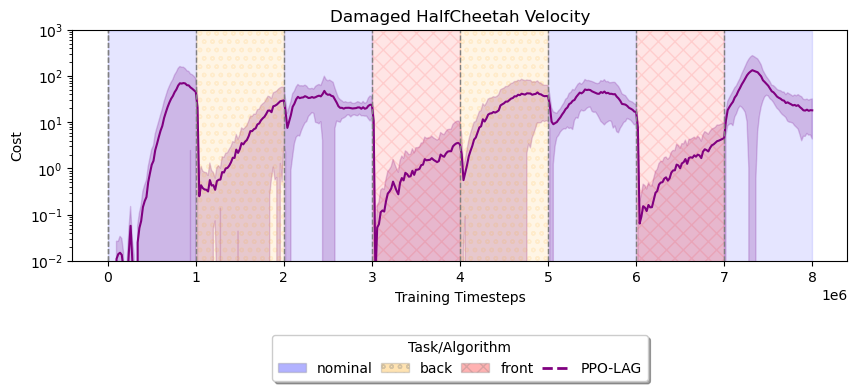}}
    \subcaptionbox{Cost for CPPO-PID on Damaged HalfCheetah Velocity.\label{fig:cppo_pid_cost}}{
        \includegraphics[width=0.48\linewidth]{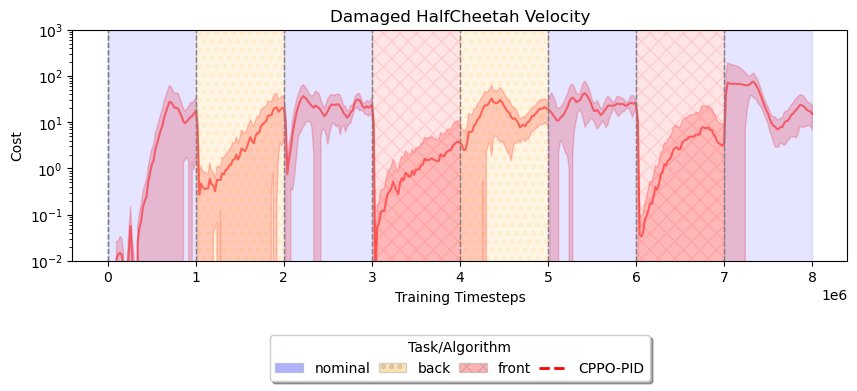}}

    \vspace{0.5cm}

    \subcaptionbox{Cost for PPO+EWC on Damaged HalfCheetah Velocity.\label{fig:ppo_ewc_cost}}{
        \includegraphics[width=0.48\linewidth]{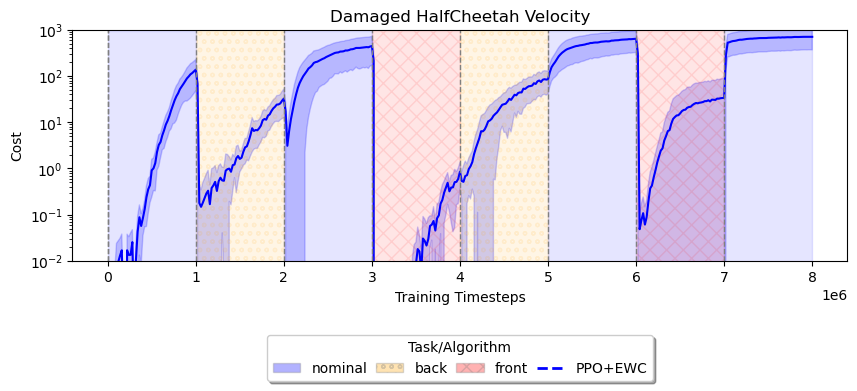}}
    \subcaptionbox{Cost for Experience Replay on Damaged HalfCheetah Velocity.\label{fig:exp_replay_cost}}{
        \includegraphics[width=0.48\linewidth]{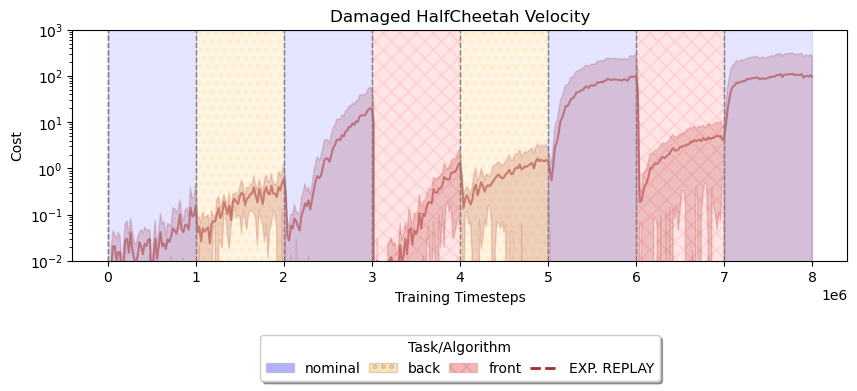}}

    \vspace{0.5cm}

    \subcaptionbox{Cost for CF-EWC on Damaged HalfCheetah Velocity.\label{fig:safe_ewc_cost}}{
        \includegraphics[width=0.48\linewidth]{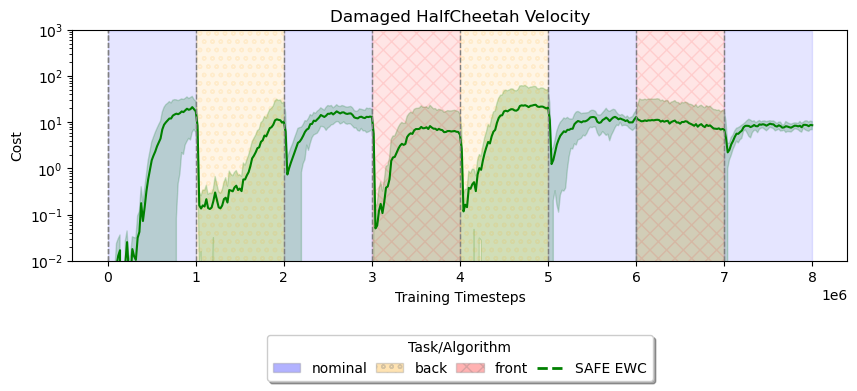}}
    \subcaptionbox{Cost for CF-EWC on Damaged HalfCheetah Velocity.\label{fig:safe_ewc_cost}}{
        \includegraphics[width=0.48\linewidth]{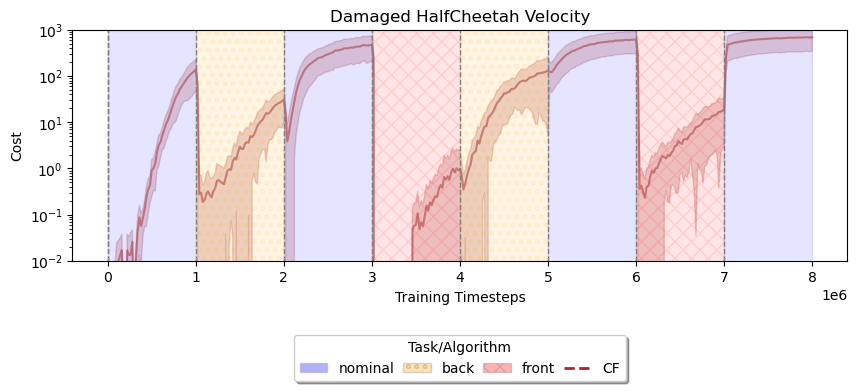}}

    \caption{Cost curves for different algorithms across 5 random seeds on Damaged HalfCheetah Velocity task.}
    \label{fig:all_costs}
\end{figure}

\begin{figure}[H]
    \centering
    \captionsetup{justification=centering}
    \subcaptionbox{Reward for PPO on Damaged Ant Velocity.\label{fig:ppo_reward}}{
        \includegraphics[width=0.48\linewidth]{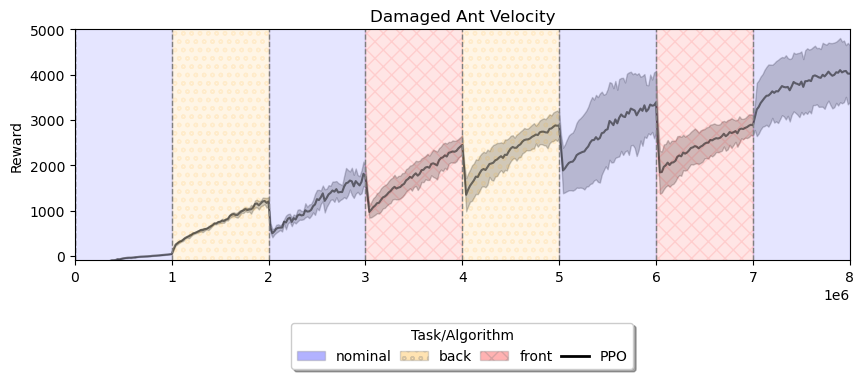}}
    \subcaptionbox{Reward for CPO on Damaged Ant Velocity.\label{fig:cpo_reward}}{
        \includegraphics[width=0.48\linewidth]{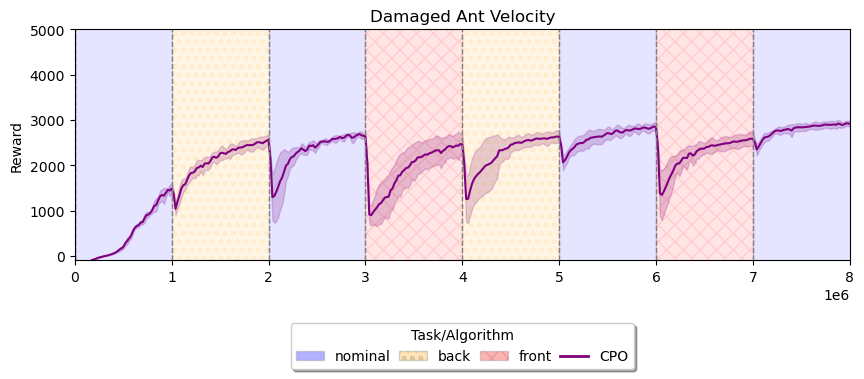}}

    \vspace{0.5cm}

    \subcaptionbox{Reward for PPO-Lag on Damaged Ant Velocity.\label{fig:ppo_lag_reward}}{
        \includegraphics[width=0.48\linewidth]{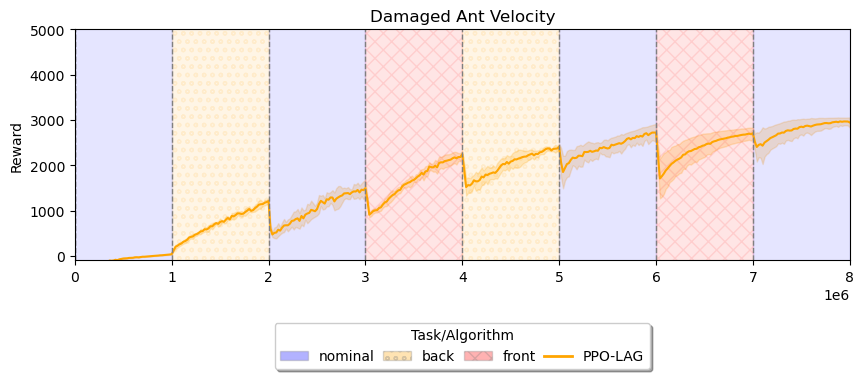}}
    \subcaptionbox{Reward for CPPO-PID on Damaged Ant Velocity.\label{fig:cppo_pid_reward}}{
        \includegraphics[width=0.48\linewidth]{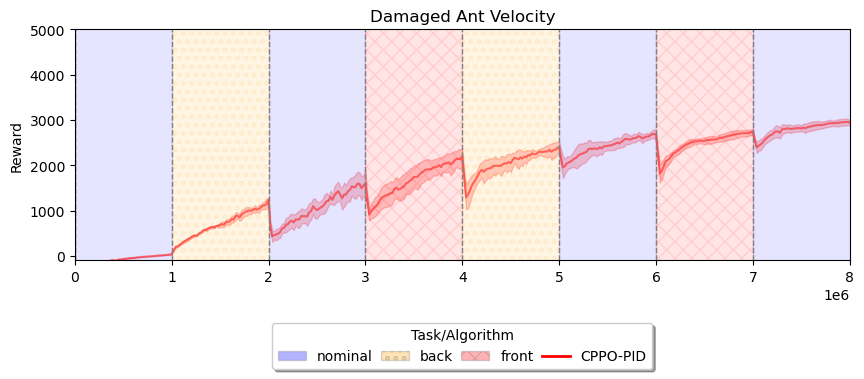}}

    \vspace{0.5cm}

    \subcaptionbox{Reward for PPO+EWC on Damaged Ant Velocity.\label{fig:ppo_ewc_reward}}{
        \includegraphics[width=0.48\linewidth]{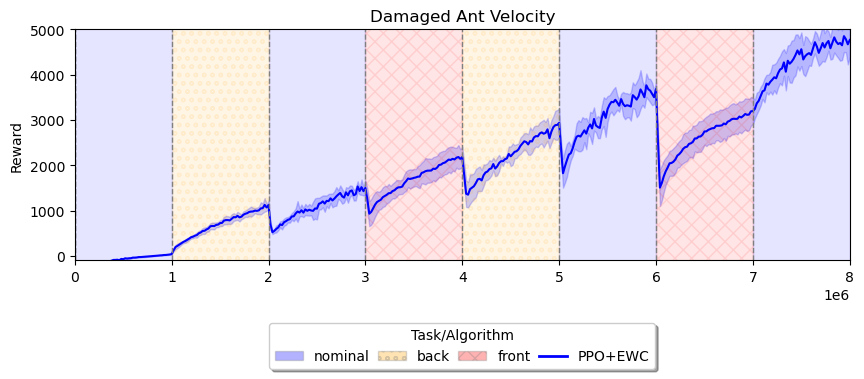}}
    \subcaptionbox{Reward for Safe EWC on Damaged Ant Velocity.\label{fig:safe_ewc_reward}}{
        \includegraphics[width=0.48\linewidth]{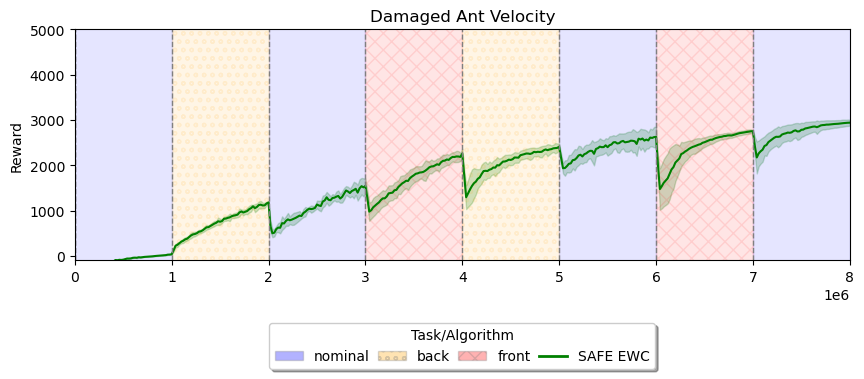}}

    \vspace{0.5cm}

    \subcaptionbox{Reward for CF-EWC on Damaged Ant Velocity.\label{fig:cf_ewc_reward}}{
        \includegraphics[width=0.48\linewidth]{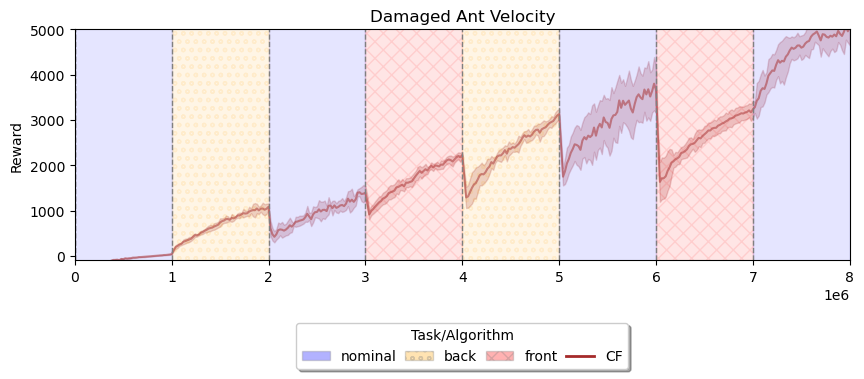}}

    \caption{Reward curves for different algorithms across 5 random seeds on Damaged Ant Velocity task.}
    \label{fig:all_rewards}
\end{figure}

\begin{figure}[H]
    \centering
    \captionsetup{justification=centering}
    \subcaptionbox{Cost for PPO on Damaged Ant Velocity.\label{fig:ppo_cost}}{
        \includegraphics[width=0.48\linewidth]{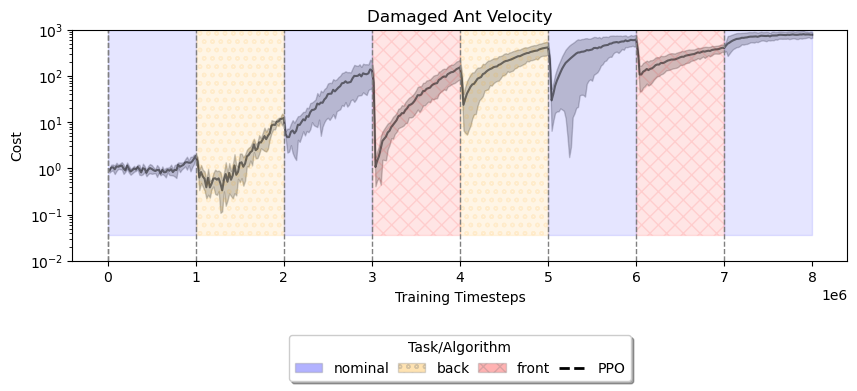}}
    \subcaptionbox{Cost for CPO on Damaged Ant Velocity.\label{fig:cpo_cost}}{
        \includegraphics[width=0.48\linewidth]{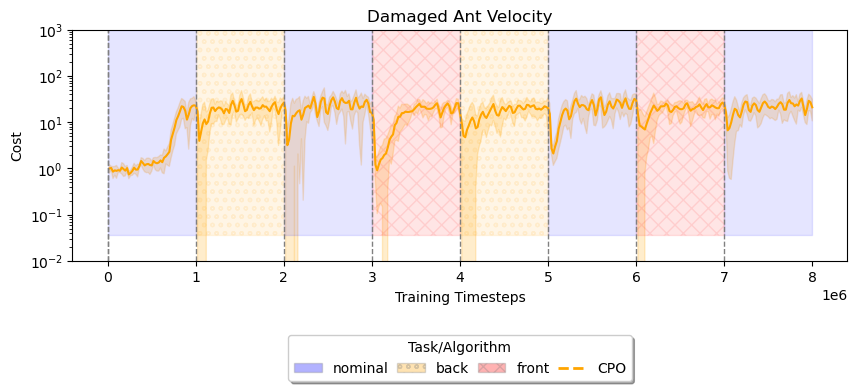}}

    \vspace{0.5cm}

    \subcaptionbox{Cost for PPO-Lag on Damaged Ant Velocity.\label{fig:ppo_lag_cost}}{
        \includegraphics[width=0.48\linewidth]{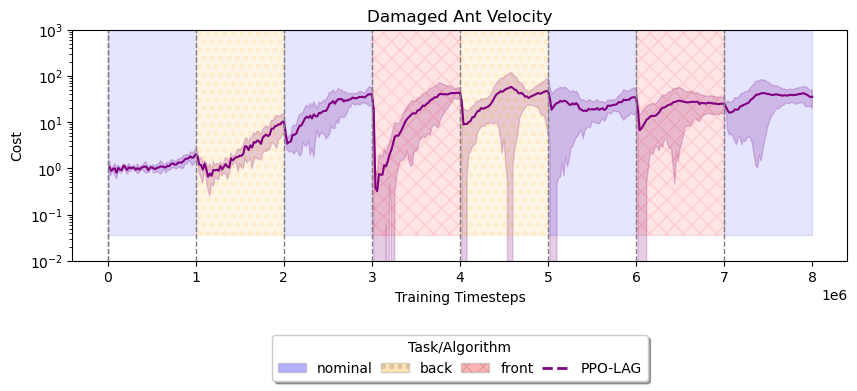}}
    \subcaptionbox{Cost for CPPO-PID on Damaged Ant Velocity.\label{fig:cppo_pid_cost}}{
        \includegraphics[width=0.48\linewidth]{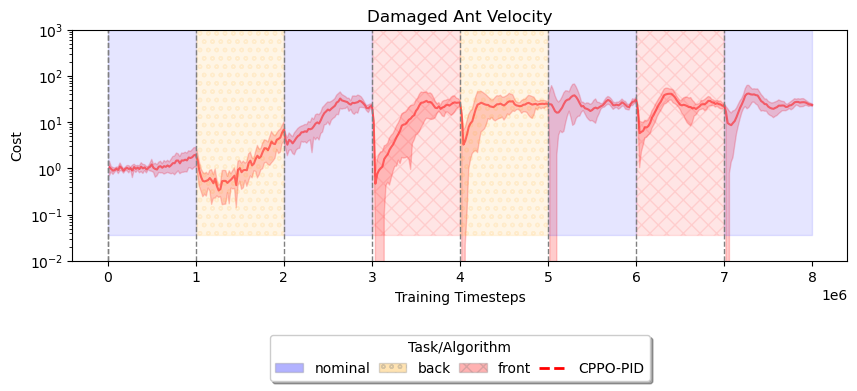}}

    \vspace{0.5cm}

    \subcaptionbox{Cost for PPO+EWC on Damaged Ant Velocity.\label{fig:ppo_ewc_cost}}{
        \includegraphics[width=0.48\linewidth]{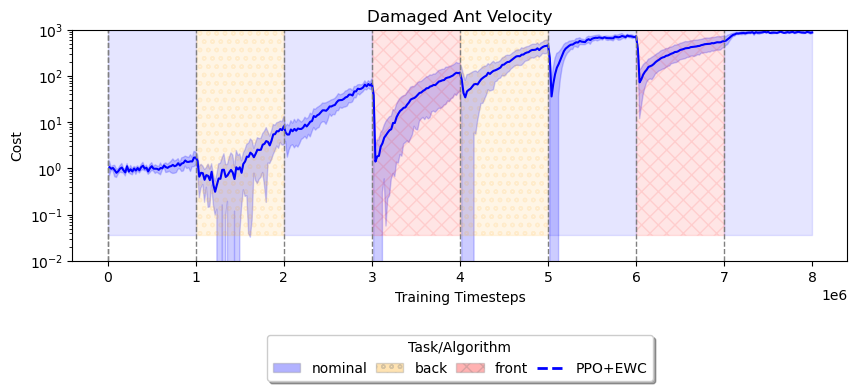}}
    \subcaptionbox{Cost for CF-EWC on Damaged Ant Velocity.\label{fig:safe_ewc_cost}}{
        \includegraphics[width=0.48\linewidth]{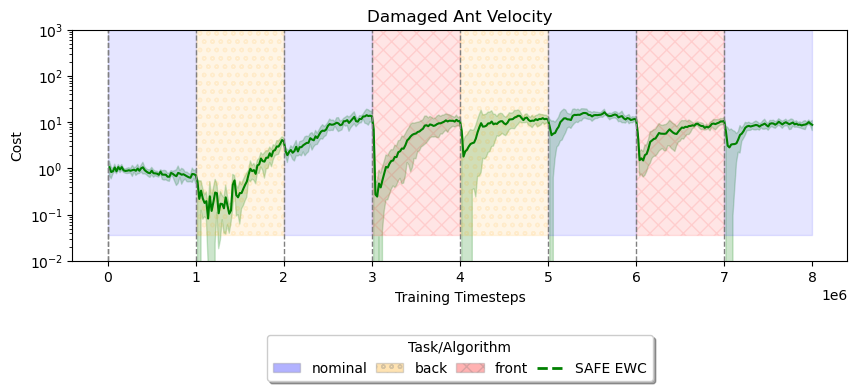}}

    \vspace{0.5cm}

    \subcaptionbox{Cost for CF-EWC on Damaged Ant Velocity.\label{fig:safe_ewc_cost}}{
        \includegraphics[width=0.48\linewidth]{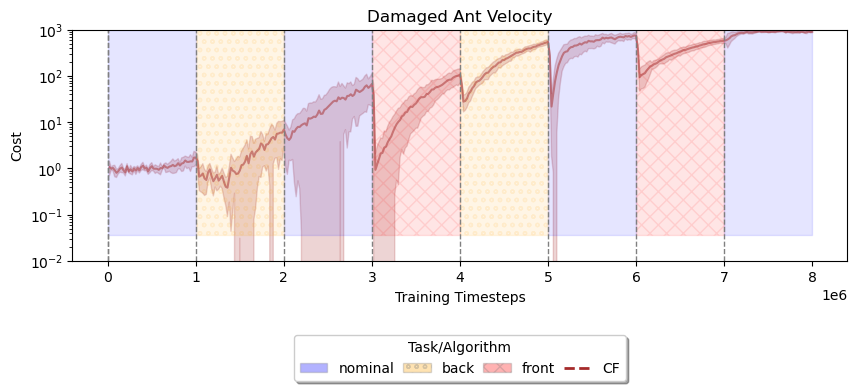}}

    \caption{Cost curves for different algorithms across 5 random seeds on Damaged Ant Velocity task.}
    \label{fig:all_costs}
\end{figure}

\begin{figure}[H]
    \centering
    \captionsetup{justification=centering}
    \subcaptionbox{Success for PPO on Safe Continual World.\label{fig:ppo_success_cw}}{
        \includegraphics[width=0.48\linewidth]{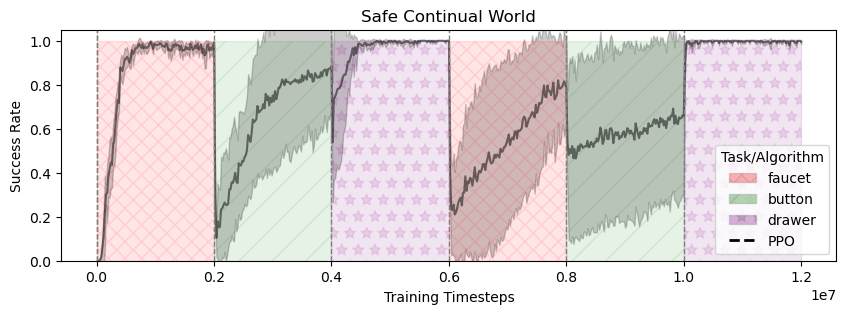}}
    \subcaptionbox{Success for CPO on Safe Continual Worldy.\label{fig:cpo_success}}{
        \includegraphics[width=0.48\linewidth]{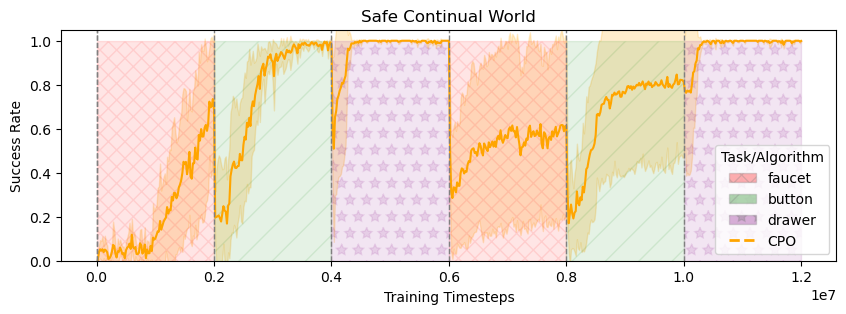}}

    \vspace{0.5cm}

    \subcaptionbox{Success for PPO-Lag on Safe Continual World.\label{fig:ppo_lag_success}}{
        \includegraphics[width=0.48\linewidth]{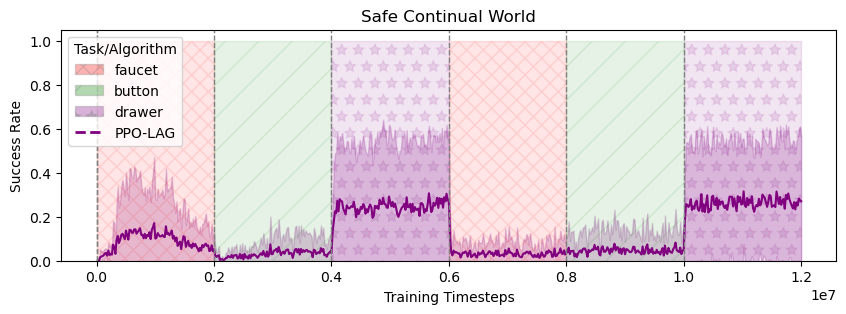}}
    \subcaptionbox{Success for CPPO-PID on Safe Continual World.\label{fig:cppo_pid_success}}{
        \includegraphics[width=0.48\linewidth]{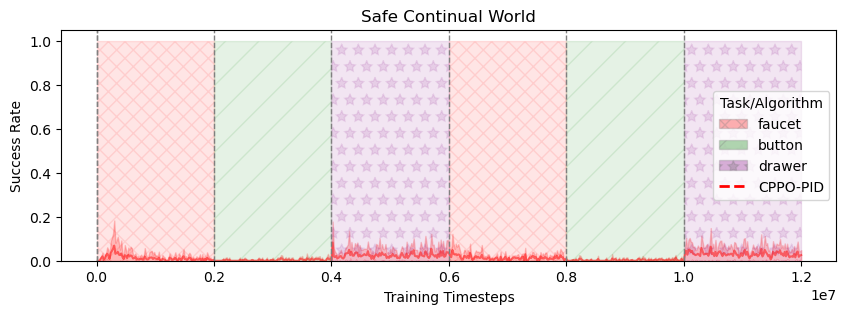}}

    \vspace{0.5cm}

    \subcaptionbox{Success for PPO+EWC on Safe Continual World.\label{fig:ppo_ewc_success}}{
        \includegraphics[width=0.48\linewidth]{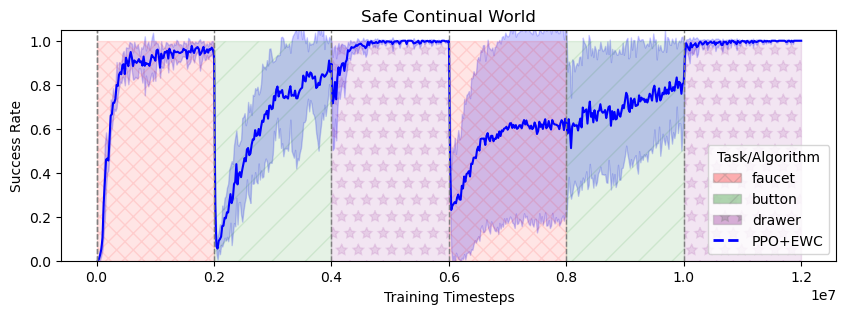}}
    \subcaptionbox{Success for Safe EWC on Safe Continual World.\label{fig:safe_ewc_success}}{
        \includegraphics[width=0.48\linewidth]{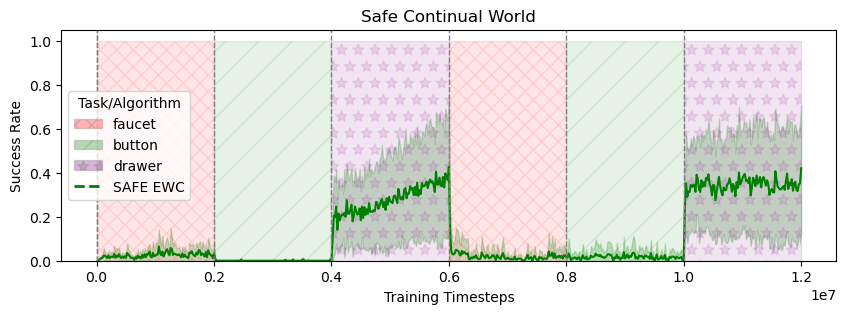}}

    \vspace{0.5cm}
    \subcaptionbox{Success for CF-EWC on Safe Continual World.\label{fig:cf_success}}{
        \includegraphics[width=0.48\linewidth]{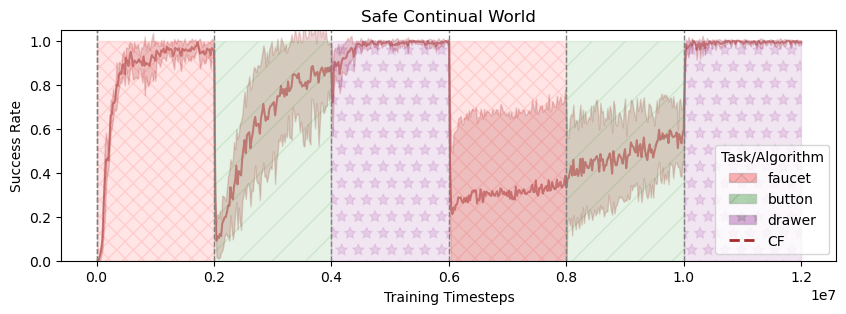}}

    \caption{Success curves for different algorithms on Safe Continual World task.}
    \label{fig:all_cw_success}
\end{figure}

\begin{figure}[H]
    \centering
    \captionsetup{justification=centering}
    \subcaptionbox{Cost for PPO on Safe Continual World.\label{fig:ppo_cost_cw}}{
        \includegraphics[width=0.48\linewidth]{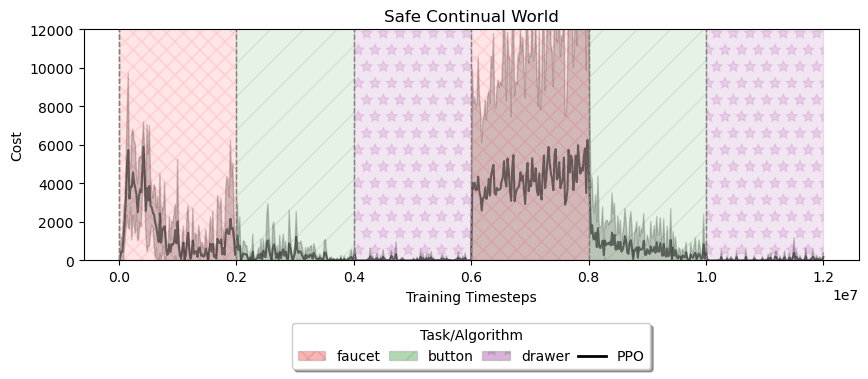}}
    \subcaptionbox{Cost for CPO on Safe Continual World.\label{fig:cpo_cost_cw}}{
        \includegraphics[width=0.48\linewidth]{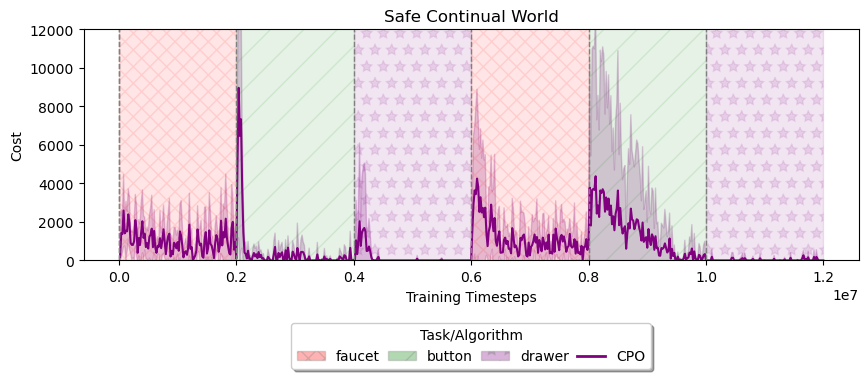}}

    \vspace{0.5cm}

    \subcaptionbox{Cost for PPO-Lag on Safe Continual World.\label{fig:ppo_lag_cost_cw}}{
        \includegraphics[width=0.48\linewidth]{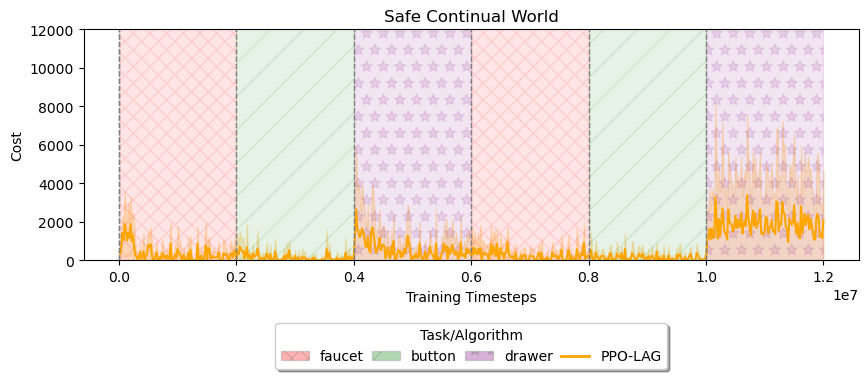}}
    \subcaptionbox{Cost for CPPO-PID on Safe Continual World.\label{fig:cppo_pid_cost_cw}}{
        \includegraphics[width=0.48\linewidth]{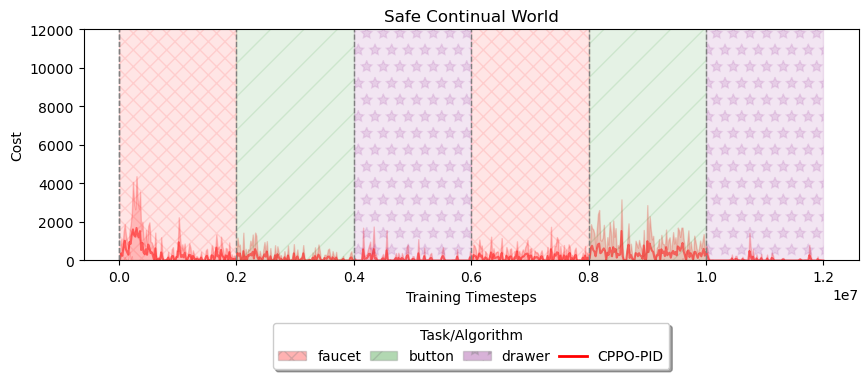}}

    \vspace{0.5cm}

    \subcaptionbox{Cost for PPO+EWC on Safe Continual World.\label{fig:ppo_ewc_cost_cw}}{
        \includegraphics[width=0.48\linewidth]{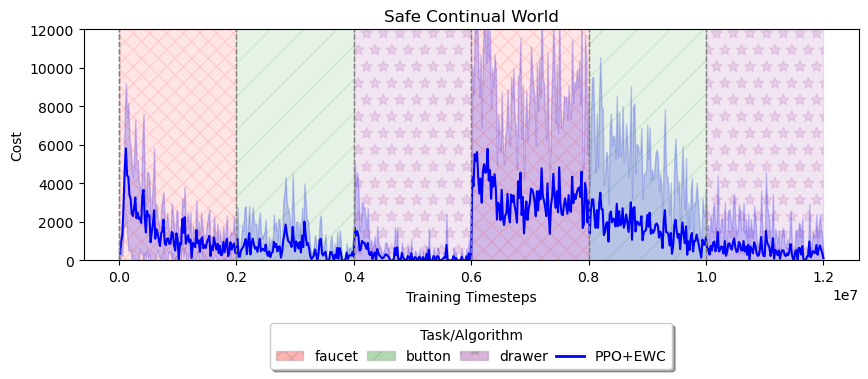}}
    \subcaptionbox{Cost for Safe EWC on Safe Continual World.\label{fig:exp_replay_cost_cw}}{
        \includegraphics[width=0.48\linewidth]{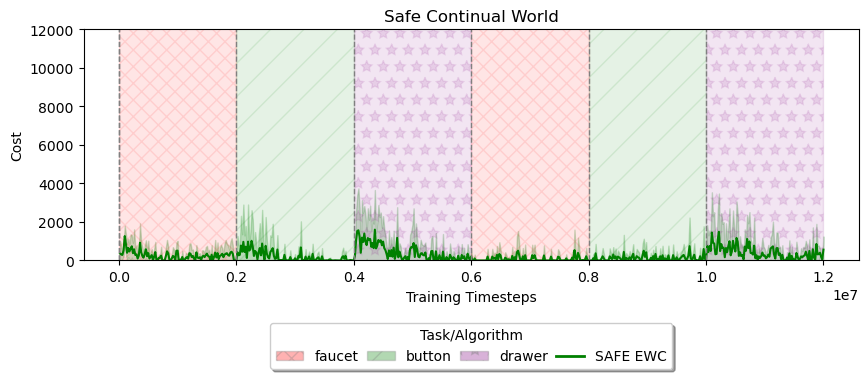}}

    \vspace{0.5cm}
    \subcaptionbox{Cost for CF-EWC on Safe Continual World.\label{fig:exp_replay_cost_cw}}{
        \includegraphics[width=0.48\linewidth]{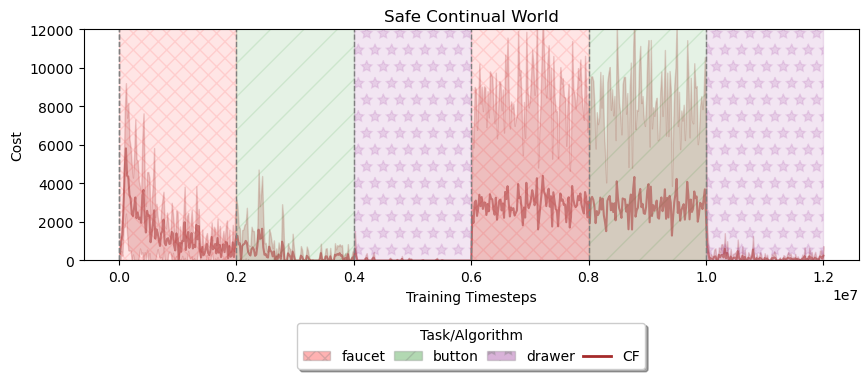}}

    \caption{Cost curves for different algorithms on Safe Continual World task.}
    \label{fig:all_cw_costs}
\end{figure}

\subsection{Combined Training Curves} \label{sec:appendix_full_curves}

In this section, we show the full training curves of each algorithm and environment on the same plots. This allows for a direct comparison of learning behaviors in safe, continual environments. For isolated training curves, which make it easier to pick out the individual algorithms, please see the previous section.

\begin{figure}[H]
    \centering
    \includegraphics[width=\linewidth]{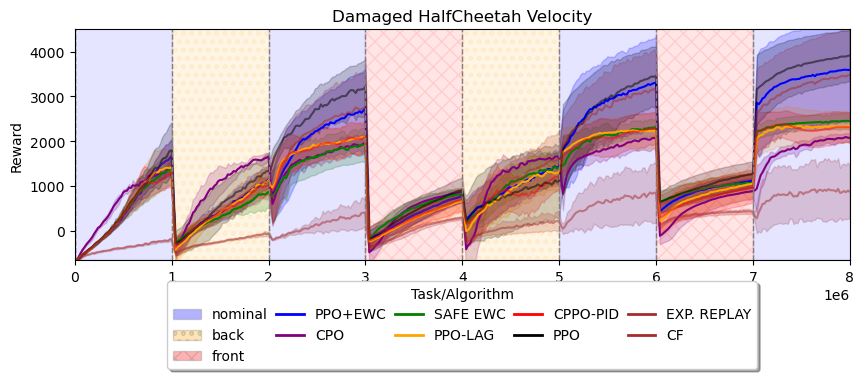}
    \caption{Reward training curves for 5 seeds of all algorithms on Damaged HalfCheetah Velocity.}
    \label{fig:all_reward_cheetah}
\end{figure}

\begin{figure}[H]
    \centering
    \includegraphics[width=\linewidth]{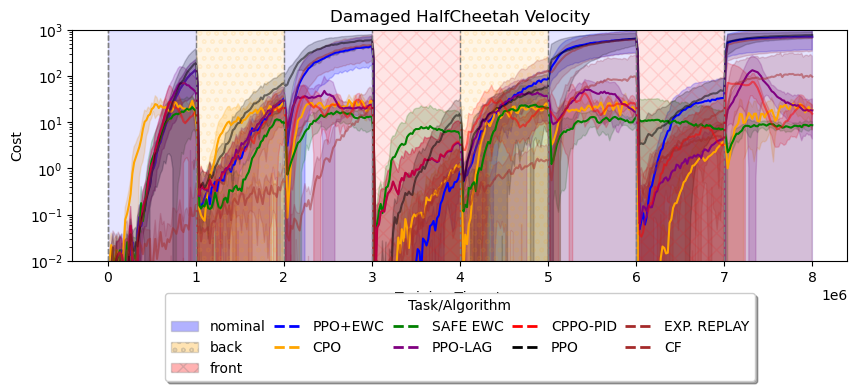}
    \caption{Cost training curves for 5 seeds of all algorithms on Damaged HalfCheetah Velocity.}
    \label{fig:all_cost_cheetah}
\end{figure}

\begin{figure}[H]
    \centering
    \includegraphics[width=\linewidth]{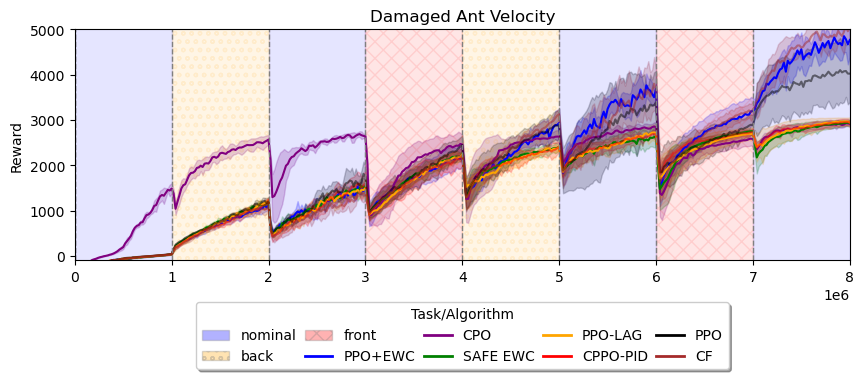}
    \caption{Reward training curves for 5 seeds of all algorithms on Damaged Ant Velocity.}
    \label{fig:all_reward_ant}
\end{figure}

\begin{figure}[H]
    \centering
    \includegraphics[width=\linewidth]{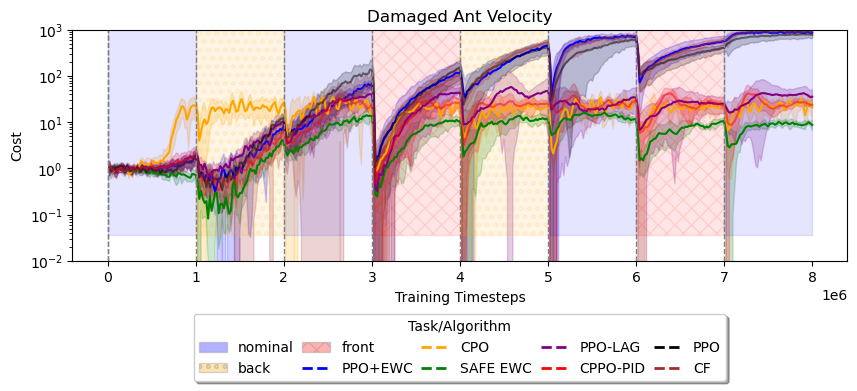}
    \caption{Cost training curves for 5 seeds of all algorithms on Damaged Ant Velocity.}
    \label{fig:all_cost_ant}
\end{figure}

\begin{figure}[H]
    \centering
    \includegraphics[width=\linewidth]{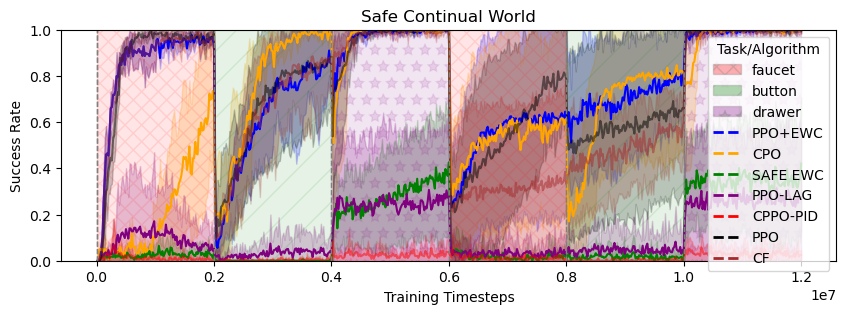}
    \caption{Success curves during training all algorithms on Safe Continual World.}
    \label{fig:all_success_cw}
\end{figure}

\begin{figure}[H]
    \centering
    \includegraphics[width=\linewidth]{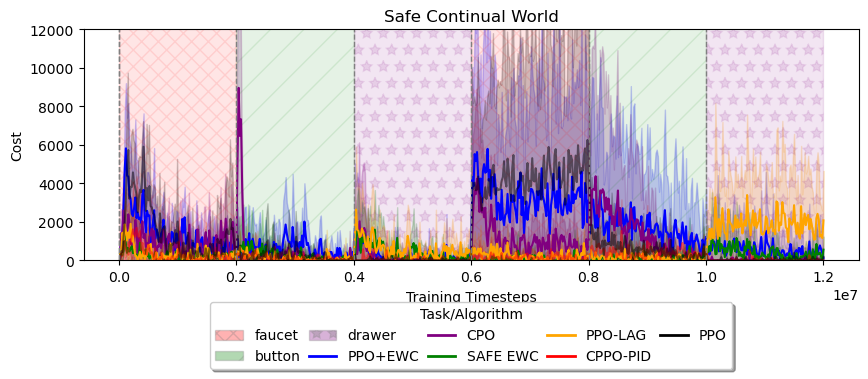}
    \caption{Cost training curves for all algorithms on Safe Continual World.}
    \label{fig:all_cost_cw}
\end{figure}

\end{document}